\documentclass[letterpaper, 10pt, journal, twoside]{IEEEtran}

\IEEEoverridecommandlockouts
\makeatletter

\let\proof\@undefined
\let\endproof\@undefined
\makeatother

\usepackage{amsmath}  
\usepackage{amssymb}  
\usepackage{amsthm}
\usepackage{amsfonts}
\usepackage{mathtools}

\usepackage{paralist}
\usepackage{fancyhdr}

\theoremstyle{definition}

\theoremstyle{remark}

\usepackage{verbatim}

\usepackage{color}
\usepackage{graphicx}
\usepackage{caption}

\usepackage{times}

\usepackage[ruled,vlined,linesnumbered]{algorithm2e}
\usepackage[noend]{algorithmic}

\usepackage[us]{datetime}
\usepackage{caption}
\usepackage[usenames,dvipsnames,table]{xcolor}

\newcommand{\threeD}{\mbox{3-D} }

\usepackage{soul}

\usepackage{todonotes}
\usepackage{subfig}

\usepackage{latexsym}
\usepackage{color}

\usepackage[noadjust]{cite}

\usepackage{caption}
\usepackage{float}

\usepackage{tikz}
\usetikzlibrary{shapes,fit,arrows,calc,intersections,
positioning,backgrounds,decorations.markings, patterns,external}

\tikzset{external/system call={latex \tikzexternalcheckshellescape -halt-on-error
    -interaction=batchmode -jobname "\image" "\texsource";
    dvips -o "\image".eps "\image".dvi;
ps2eps "\image.eps"}}

\let\labelindent\relax 
\usepackage[inline]{enumitem} 
\setenumerate[0]{label=\roman*.}

\usepackage{subfiles}
\usepackage{bm}

\tikzset{
  connect/.style args={(#1) to (#2) over (#3) by #4}{
    insert path={
      let \p1=($(#1)-(#3)$), \n1={veclen(\x1,\y1)},
      \n2={atan2(\x1,\y1)}, \n3={abs(#4)}, \n4={#4>0 ?180:-180}  in
      (#1) -- ($(#1)!\n1-\n3!(#3)$)
      arc (\n2:\n2+\n4:\n3) -- (#2)
    }
  },
}

\usepackage{scalerel}
\usetikzlibrary{svg.path}
\definecolor{orcidlogocol}{HTML}{A6CE39}
\tikzset{
  orcidlogo/.pic={
    \fill[orcidlogocol] svg{M256,128c0,70.7-57.3,128-128,128C57.3,256,0,198.7,0,128C0,57.3,57.3,0,128,0C198.7,0,256,57.3,256,128z};
    \fill[white] svg{M86.3,186.2H70.9V79.1h15.4v48.4V186.2z}
    svg{M108.9,79.1h41.6c39.6,0,57,28.3,57,53.6c0,27.5-21.5,53.6-56.8,53.6h-41.8V79.1z M124.3,172.4h24.5c34.9,0,42.9-26.5,42.9-39.7c0-21.5-13.7-39.7-43.7-39.7h-23.7V172.4z}
    svg{M88.7,56.8c0,5.5-4.5,10.1-10.1,10.1c-5.6,0-10.1-4.6-10.1-10.1c0-5.6,4.5-10.1,10.1-10.1C84.2,46.7,88.7,51.3,88.7,56.8z};
  }
}
\newcommand\orcidicon[1]{\href{https://orcid.org/#1}{\mbox{\scalerel*{
        \begin{tikzpicture}[yscale=-1,transform shape]
          \pic{orcidlogo};
        \end{tikzpicture}
}{|}}}}

\makeatletter
\let\NAT@parse\undefined
\makeatother

\usepackage{hyperref} 
\hypersetup{
  colorlinks,
  citecolor=black,
  filecolor=black,
  linkcolor=black,
  urlcolor=black,
  pdfauthor={},
  pdfsubject={},
  pdftitle={}
}

\setlist[itemize]{noitemsep, nosep}

\usepackage{siunitx}
\usepackage{ifthen}

\usepackage{booktabs}

\renewcommand{\baselinestretch}{0.99}
\usepackage{tablefootnote}

\makeatletter
\def  \input@path{{./../fig/},{./fig/}}
\makeatother

\title{%
  Dronument: System for Reliable Deployment of Micro Aerial Vehicles in Dark Areas of\\Large Historical Monuments
}

\author{
  Pavel Petr\'{a}\v{c}ek$^{\orcidicon{0000-0002-0887-9430}}$,
  V\'{i}t Kr\'{a}tk\'{y}$^{\orcidicon{0000-0002-1914-742X}}$, and
  Martin Saska$^{\orcidicon{0000-0001-7106-3816}}$%
  \thanks{Manuscript received: September 10, 2019; Revised December 11, 2019; Accepted January 9, 2020.}
  \thanks{This paper was recommended for publication by Editor Jonathan Roberts upon evaluation of the Associate Editor and Reviewers' comments.
  This work was supported by project no. DG18P02OVV069 in program NAKI II, by CTU grant no. SGS17/187/OHK3/3T/13, and by the Grant Agency of the Czech Republic under grant no. 17-16900Y.}
  \thanks{The authors are with the Faculty of Electrical Engineering, Czech Technical University in Prague, 166 36 Prague 6, Czech Republic {\tt\footnotesize\{\href{mailto:pavel.petracek@fel.cvut.cz}{pavel.petracek}|\href{mailto:vit.kratky@fel.cvut.cz}{vit.kratky}|\href{mailto:martin.saska@fel.cvut.cz}{martin.saska}\}@fel.cvut.cz}.}
  \thanks{Digital Object Identifier (DOI): see top of this page.}
}

\begin{document}

\newcommand{\PREPRINTYEAR}{2020}
\newcommand{\PREPRINTPUBLISHER}{IEEE Robotics and Automation Letters}
\newcommand{\DOI}{10.1109/LRA.2020.2969935}

\markboth{\PREPRINTPUBLISHER, \PREPRINTYEAR. DOI: \DOI}{}

\fancyhead{}
\chead{©\PREPRINTPUBLISHER, \PREPRINTYEAR. DOI: \DOI}
\pagestyle{fancy}
\thispagestyle{plain}

\onecolumn
\pagenumbering{gobble}
{
  \topskip0pt
  \vspace*{\fill}
  \centering
  \LARGE{%
    © \PREPRINTYEAR~\PREPRINTPUBLISHER\\\vspace{1cm}
    Personal use of this material is permitted.
    Permission from \PREPRINTPUBLISHER~must be obtained for all other uses, in any current or future media, including reprinting or republishing this material for advertising or promotional purposes, creating new collective works, for resale or redistribution to servers or lists, or reuse of any copyrighted component of this work in other works.\\\vspace*{1cm}DOI: \DOI}
    \vspace*{\fill}

}
\twocolumn 
\pagenumbering{arabic}

\maketitle

\begin{abstract}
  This letter presents a self-contained system for robust deployment of autonomous aerial vehicles in environments without access to global navigation systems and with limited lighting conditions.
  The proposed system, application-tailored for documentation in dark areas of large historical monuments, uses a unique and reliable aerial platform with a multi-modal lightweight sensory setup to acquire data in human-restricted areas with adverse lighting conditions, especially in areas that are high above the ground.
  The introduced localization method relies on an easy-to-obtain \threeD point cloud of a historical building, while it copes with a lack of visible light by fusing active laser-based sensors.
  The approach does not rely on any external localization, or on a preset motion-capture system.
  This enables fast deployment in the interiors of investigated structures while being computationally undemanding enough to process data online, onboard an MAV equipped with ordinary processing resources.

  The reliability of the system is analyzed, is quantitatively evaluated on a set of aerial trajectories performed inside a real-world church, and is deployed onto the aerial platform in the position control feedback loop to demonstrate the reliability of the system in the safety-critical application of historical monuments documentation.
\end{abstract}

\begin{IEEEkeywords}
  Aerial Systems: Applications, Aerial Systems: Perception and Autonomy, Localization
\end{IEEEkeywords}

\vspace*{-0.6em}
\section{Introduction}

\IEEEPARstart{I}{n} recent years, massive advances have been made in the technology of aerial vehicles capable of vertical landing and takeoff, in terms of control, reliability, and autonomy.
These multirotor vehicles, which we will refer to as Micro Aerial Vehicles (MAVs), have become extremely popular for their flexibility, diversity, and potential for applicability and amusement.
The broad application spectrum of MAV systems ranges from \threeD mapping and deployment in search \& rescue scenarios to wildlife \& nature conservation.

This letter presents a unique self-localization system for a specialized use of MAV teams - autonomous documentation of historical monuments - derived from the interest expressed by end-users with expertise in restoration and conservation.
The current procedure used during regular studies for restoration works requires a large scaffold to be constructed in order to monitor the condition of a building and its artifacts.
An MAV platform can supply the same documentation and inspection techniques as those provided by the experts, but in locations unreachable by people except with the use of a large and expensive scaffolding installation, or in locations which had never been documented before during an initial survey.
The MAV platform significantly speeds up the duration and significantly scales down the cost of the restoration works, and offers data acquisition from previously impossible angles and unreachable locations.

\begin{figure}[t]
  \vspace*{-0.5em}
  \hspace*{-0.4em}
  \centering
  \subfloat{
    \includegraphics[width=0.49\columnwidth]{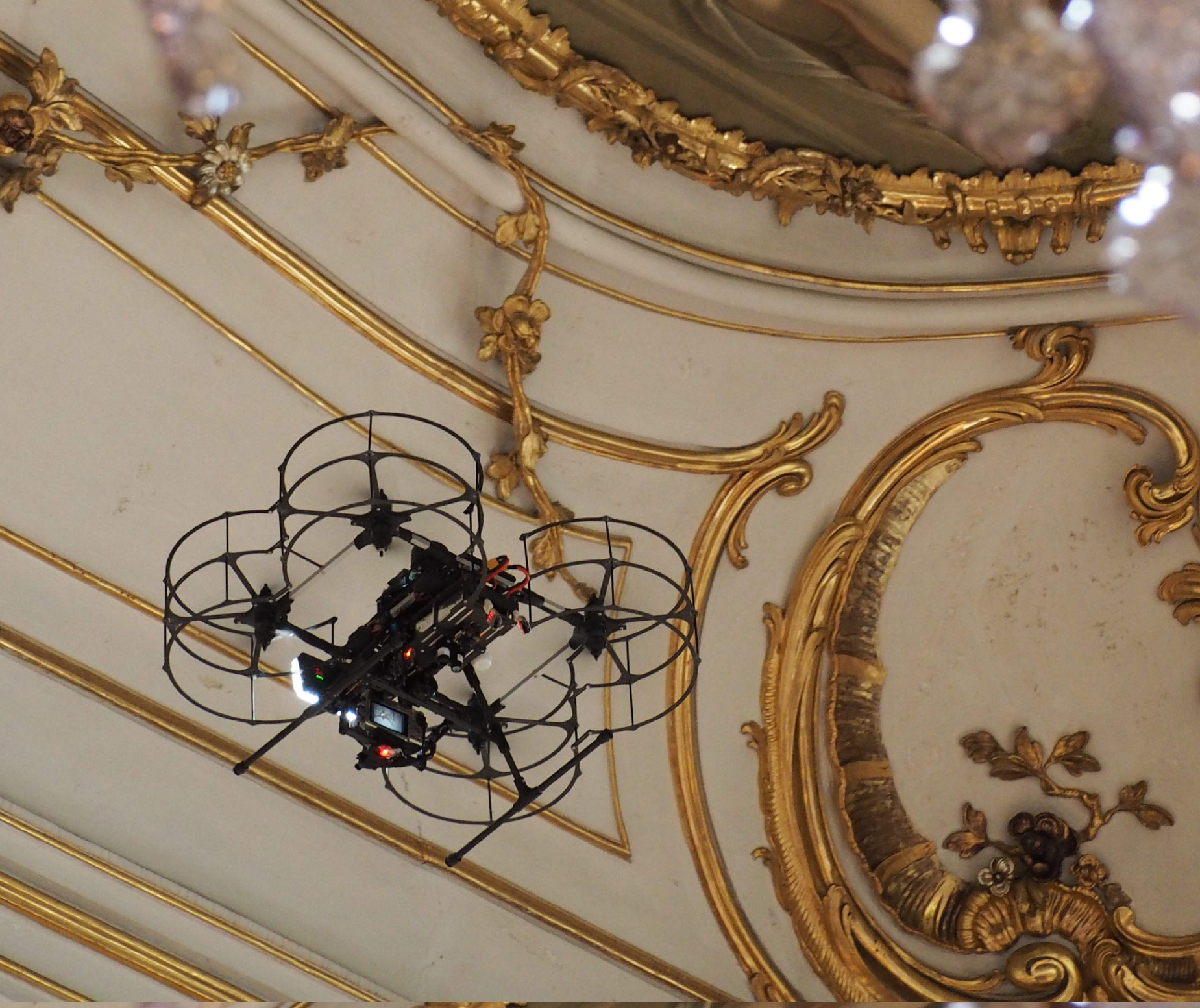}
  }
  \hspace*{-0.8em}
  \subfloat{
    \includegraphics[width=0.49\columnwidth]{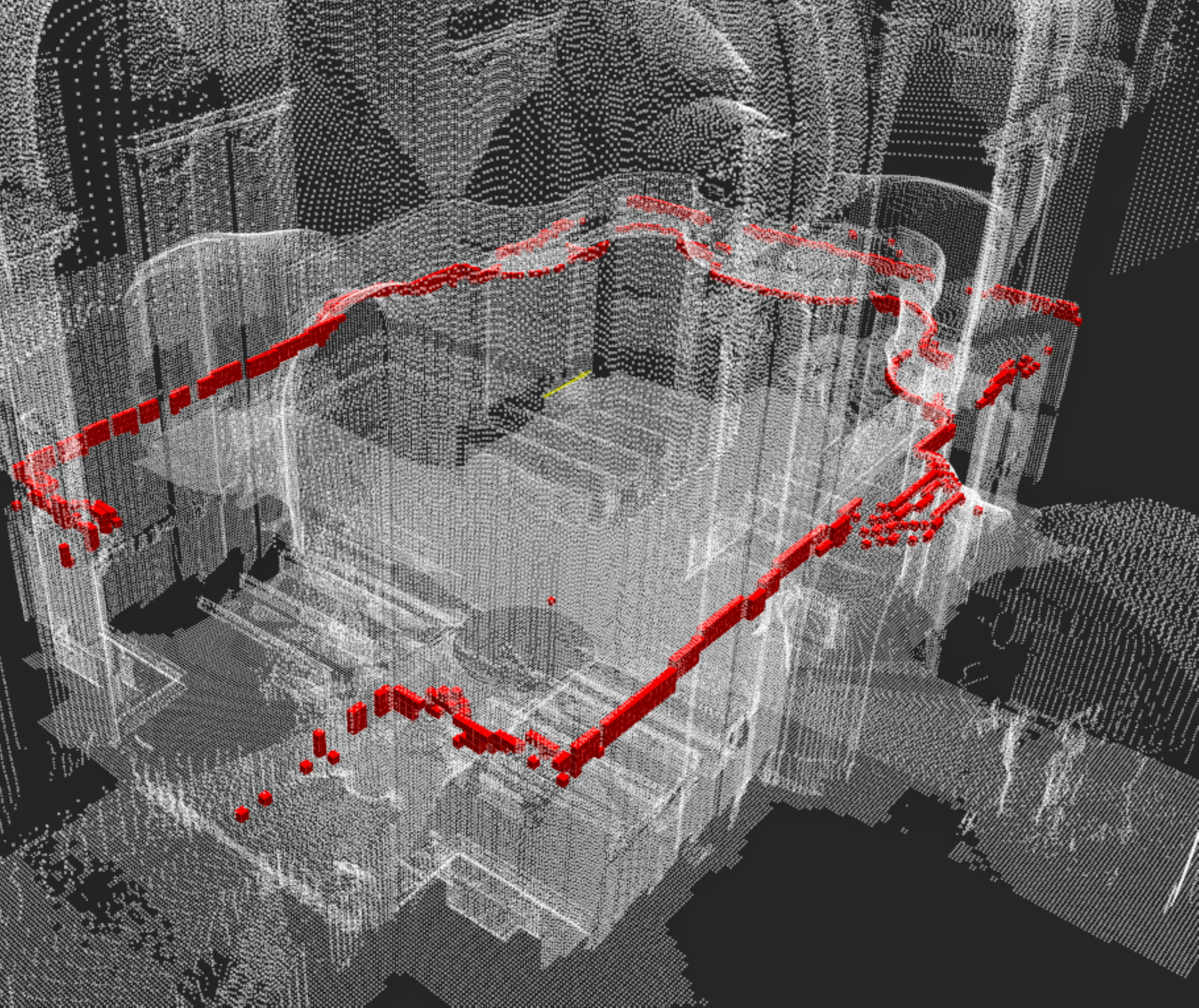}
  }
  \caption{Demonstration of the presented system during documentation of the parliamentary hall of the UNESCO castle in Krom\v{e}\v{r}\'{i}\v{z}, Czech Republic, with an example of onboard sensory data registration into the object map}
  \label{fig:motivation_title}
  \vspace*{-1.5em}
\end{figure}

The proposed system is designed for deployment in historical monuments, such as ancient or modern, war-damaged, dilapidated or restored cathedrals, chapels, churches, mausoleums, castles, and temples with dimensions varying from small chapels up to large cathedrals.
The deployment of robots in these operational environments is a challenging task due to the absence of a global navigation satellite system (GNSS), the adverse lighting conditions, and numerous other challenges summarized later on in {\autoref{sec:experience_gained}}.
An aerial system that handles all the challenges has to provide exceptional robustness, which we propose to achieve by introducing a precise model-based control approach, reliable real-time state estimation, a high level of sensor \& actuator redundancy, and feasible mission planning \& navigation.
In this letter, we also address in detail the problem of real-time state estimation acting as a state observer for MAV control and mission navigation modules in the tackled \mbox{\textit{GNSS-denied}} environments of large historical buildings.
The proposed system relies on a lightweight sensory setup composed of a 2-D laser-scanner and two point-distance rangefinders, and on a map of a historical site pre-generated in the form of a \mbox{\threeD} point cloud provided by a terrestrial laser scanner (TLS).
Our laser-inertial approach to indoor localization fuses an onboard IMU and a locally refined global state estimation, while it processes the data onboard an MAV and estimates the global state in real-time.



\subsection{Related Work}
Until now, documentation of interiors and exteriors of buildings and facilities has generally been performed manually by generating a \threeD site model using a TLS, together with scan registration post-processing \cite{7548040} or \mbox{photogrammetry \cite{laquila_documentation}} requiring geo-reference information.
The emergence of automation procedures has speeded up the scanning processes.
Examples are the \textit{Zebedee} \cite{ZLOT2014670} and the \textit{LIPS} \cite{8594463} hand-held mobile \threeD laser mapping systems, and even ground robot mapping approaches \cite{BORRMANN2015105}.

Using aerial systems introduces the advantage of rapid documentation even in human-unreachable locations.
These systems are being employed for documentation purposes \cite{drones2010002, laquila_documentation}, but most of them are deployed manually outdoors and require GNSS to obtain geo-referenced data.

Although the lack of GNSS can be bypassed by exploiting a preset external localization system capable of a high accuracy localization, this approach is not scalable for documentation of large structures with limited access time.
Other approaches for \textit{GNSS-denied} localization include visual-inertial simultaneous localization and mapping (VI-SLAM), whose recent advances are well described in \mbox{{\cite{ABOUZAHIR201814}} and {\cite{Huang2019ICRA}}}.
A mono- or stereo-camera SLAM is a thoroughly studied problem for an MAV, due to the lightweight of ubiquitous cameras.
One example of a system that attempts to integrate online SLAM is the Open Vision Computer \cite{DBLP:journals/corr/abs-1809-07674}, which is an embedded off-the-shelf FPGA module that handles a visual SLAM independent of other onboard subsystems.
The state-of-the-art vision-based ORB-SLAM2 \cite{7946260} was tested in real conditions with lighting conditions similar to the desired environments.
However, it was found to be ineffective, and it was disregarded for reasons described in \autoref{sec:experience_gained}.
The lighting issues motivated the development of the system presented here, which works under the specified conditions.\looseness=-1

The authors in \mbox{\cite{chimneyspector, 7914761, 8392775}} have presented applications, which share a considerable number of common characteristics with indoor documentation of historical structures.
A laser-inertial system for inspecting chimneys is presented in {\cite{chimneyspector}}, while a laser-visual-inertial system for inspecting penstocks and tunnels is presented in {\cite{7914761}}.
In comparison to our application, environments tackled in {\cite{chimneyspector}} and {\cite{7914761}} are well-structured and homogeneous for onboard sensors, which makes it possible to tune the system for these specific conditions.
On the contrary, our task requires a much higher level of complexity.
The authors in {\cite{8392775}} focused on inventory applications in warehouses.
Their laser-visual-inertial setup is suited for fast flights in complex dynamic environments in order to speed up a periodic inventory audit.
However, our targeted scenario requires minimalist MAV dimensions, and slower and more accurate operation with respect to a variable onboard payload.
In contrast to our lightweight sensory setup, the systems in \mbox{\cite{chimneyspector, 7914761, 8392775}} employ a heavyweight \mbox{\threeD} lidar.
Moreover, \mbox{the systems in {\cite{7914761}} and {\cite{8392775}} fuse} visual information from a set of onboard cameras, which is not suitable for the tackled environments with \mbox{adverse lighting conditions.}%


Apart from that, only one work using MAV in the context of documentation of historical buildings has been found \cite{slam_budapest}.
This work evaluates the performance of several SLAM and SFM methods during a \mbox{3-D} model reconstruction of a single historical site.
However, the authors of {\cite{slam_budapest}} perform only an offline evaluation of their methods on an outdoor aerial trajectory and do not deploy these methods in \textit{GNSS-denied} environments nor in a position control loop of an MAV.%

Documentation systems often extend their applications with a TLS to assist with the digital preservation of the scanned sites \cite{7433210}, which we likewise propose in our system architecture to boost the robustness of the system.
In \cite{PerezGrau2018AnAF}, map-based \threeD Monte Carlo localization (MCL) using an RGB-D camera provides a global state estimate.
In contrast to this manuscript, our method utilizes a 2-D scanner instead of an RGB-D camera.
This provides planar \SI{360}{\degree} information, making it independent from orientation.
Our method goes further by refining the global estimate on a local map by a scan matching technique to yield faster and more accurate results.
The authors of manuscripts \cite{s17061268, wang_wang} fuse scan matching output, IMU, and a down-oriented rangefinder.
Our proposed method extends the setup with global initialization and fusion of an up-oriented rangefinder.
The importance of the up-oriented measurements rises significantly during flights over heterogeneous objects (church benches), where vertical estimate exploits the homogeneous nature of ceilings.



\subsection{Contributions}
This letter addresses problems of the deployment of aerial systems in the safety-critical application of historical monument documentation.
The stability of the system originates from carrying out tests in real-world historical objects in the course of two years of a research and culture project in close cooperation with the National Heritage Institute of the Czech Republic.
The insights into developments for real-world deployment presented here tackle the motivations and constraints of the highly challenging environments guided by end-users from outside the robotic community.
The main contributions of this letter are:
\begin{enumerate}[label=(\roman*)]
  \item It introduces a unique, highly reliable system for deployment in environments with low feature density and atrocious lighting conditions.
  \item It develops a unique hardware and software aerial platform designed in close consultation with restorers and conservationists, using experience from deployment of the system in numerous individual historical objects.
  \item It presents and shares the experience of what we believe to be the most comprehensive project in the field of autonomous documentation of historical monuments by an aerial system.
  \item It presents a robust light-independent localization system for interiors of historical buildings relying on 2-D lidar as its primary sensor.
\end{enumerate}

The \threeD localization offers precise full 6 degrees-of-freedom estimation, providing fast and robust state estimation integrated into a feedback loop of an MAV position control.
Based on a quantitative analysis evaluated on aerial ground-truth data in \autoref{sec:experimental_evaluation}, our approach reaches persistent RMSE precision below \SI{0.23}{\metre}.
The drift-free system that is presented yields greater accuracy than map-based localization for autonomous cars
\cite{WANG2017276}
and comparable accuracy to a drift-prone method \cite{DBLP:journals/corr/abs-1904-06993} employing a \threeD scanner on a ground vehicle.\looseness=-1





\section{Motivation}
\label{sec:application}
The Dronument (Drone \& Monument) project sets out to deploy MAVs for autonomous data acquisition in human-unreachable areas.
The self-contained system presented here can be deployed in three modes (manual, semi-autonomous, and fully autonomous), as allowed by the heritage institute and/or the superintendent of the structure.
These modes are specified as follows:
\begin{enumerate}[label=(\roman*)]
  \item \textit{manual:} a human operator controls all aspects of the flight using an operating transmitter, while the MAV is autonomously localized in the environment to associate gathered data with the \threeD map,
  \item \textit{semi-autonomous:} a human operator commands the flight, while onboard systems control the sensory data acquisition and provide control feedback with respect to obstacles in a \threeD neighborhood, and
  \item \textit{autonomous:} a human only specifies objects-of-interest (OoI) for documentation, and the system handles each stage of the entire mission - takeoff, stabilization \& control, localization, navigation, trajectory optimization, data acquisition, and landing.
\end{enumerate}
In addition to the deployment of a single MAV, the system is prepared for use in cooperative multi-MAV scenarios, as required for some documentation tasks.
Typical non-invasive documentation consists of a multiple spectrum survey to obtain specific information valuable for various restoration purposes.
For example, the use of different spectra contributes to more precise dating of paintings, as the glow of pigment combinations is unique to a certain period.
Examples of single (S) and cooperative (C) tasks are:

\hyphenation{photogra-phic}
\begin{itemize}
  \item \textit{Direct lighting (S)$^*$:} lighting of the scene from an onboard light with the illumination axis collinear with the optical axis of the camera.
  \item \textit{Reflectance Transformation Imaging (C):} a photographic technique for capturing the shape of a surface and the color of an object by combining photographs of the objects taken from a semi-static camera on an MAV under various illumination provided by a different MAV \cite{kratky_ral}.
  \item \textit{Three-point \& strong-side lighting (C)$^*$:} filming techniques \cite{6634193} in which 1-3 sources of light are used in different locations relative to the optical axis of the camera.
    In our previous work \cite{saska17etfa}, a Model Predictive Control (MPC) approach is proposed for controlling a formation of MAVs with respect to the lighting techniques during an aerial deployment of cooperative teams in this task.
  \item \textit{Radiography \& UV screening (C):} a method for viewing the internal structure of an object (e.g., a statue) by exposing it to X-ray or UV radiation (emission source onboard the first MAV) captured behind or in front of the object by a detector (the second MAV).
  \item \textit{\threeD reconstruction (S/C)$^*$:} a method for aggregating the shape and the appearance of an object by combining laser- and/or vision-based information into a \threeD model.
  \item \textit{Photogrammetry (S/C)$^*$:} a method for extracting measurements of real objects from photographs.
\end{itemize}
Examples of the tasks marked with ($*$) can be found within the additional multimedia materials available in {\cite{mrs_dronument}}.





\section{Experience Gained}
\label{sec:experience_gained}
Over the last two years, more than 10 objects (mainly in Moravia, Czech Republic) were documented during the ongoing development phase of the presented system.
Outputs of these documentation deployments supplied restorers and conservationists with valuable information in state assessment of multiple artifacts within the structures during the initial survey phase.
Although some of the documented structures are shown in Figures {\ref{fig:motivation_title}}, {\ref{fig:aerial_platform}}, and {\ref{fig:ground_truth_dataset}}, the complete list, together with additional multimedia materials, can be found in {\cite{mrs_dronument}}.
During the experimental phase, many lessons from the robotic as well as the restoration point of view have been learned.
The acquired experiences for objects of various sizes, shapes, and structures have influenced the system throughout the development and are herein shared.%

The indoor surveys are conducted in the close vicinity of heavy-structure buildings.
As a consequence, either GNSS is not available at all, or the system is not reliable enough, leading to \emph{GNSS-denied} operations.
In order to overcome the absence of GNSS, a local localization system must be used.
This is true even when exteriors are being documented, with the intention to document facades in their close proximity in order to capture details of artifacts from various points of view (see exteriors documentation in {\cite{mrs_dronument}}).

Insufficient \emph{lighting conditions} in a surveyed object is an impediment for two main reasons.
First, it degrades the performance of vision-based odometry and SLAM systems, which have been heavily researched over the last three decades.
Second, it lowers the quality of the photographs taken in the visible spectrum, as they require decent lighting of the scene.
These two issues motivated research in the Dronument project, leading to the development of a novel robust localization system (see \autoref{sec:state_estimation}) purposely designed for autonomous flying in specific environments of this kind.

Experience has shown that vision-based localization is limited also by \emph{feature extraction} shaped by two main characteristics.
First, it is a common occurrence to fly along protracted segments of white wall lacking any visual features at all.
Second, old religious buildings include extensive symmetric and repetitive visual patterns, such as grid flooring and artistic elements.
A common example of such an artistic element are repetitive ledge supports shown on the right side of \autoref{fig:aerial_platform}.
Together with the lighting conditions, these considerations make most vision-based systems ineffective.

The use of a prior knowledge in the form of a \emph{global map} obtained prior to the deployment of an MAV is extremely beneficial for three main reasons.
First, it facilitates the robotic problem, supplies additional robustness to the system and supports system reliability by serving as a baseline.
Second, it yields an opportunity to associate captured onboard data with a \mbox{3-D} map, which provides well-arranged data output for the end-user.
Third, the visualization of the flight plan in a \mbox{3-D} model is comprehensible for everyone - robotic experts, restorers, filmmakers - and it is necessary for confirmation purposes by an aviation authority, the heritage institute, and/or the administrator of the structure.

The proposed use of an MAV requires us to consider the \emph{diversity of environments} in which it will operate.
These environments contain distinct features - cluttered spaces, symmetric blueprints, balconies, stairs, glass windows, vault ceilings, and hanging strings.
This forced the system design to include sensor and actuator redundancy, the use of a global map, and mechanical protection of the propellers.
The test flights showed that the 2-D lidar that was used is ineffective for detecting thin obstacles, such as chandelier ropes and lighting cables.
For this specific reason, a \threeD camera is employed to detect these obstacles in front of an MAV.

Even when MAVs are deployed in historical buildings, the presence of wind gusts ascribed to the stack effect (opened windows, doors) is non-negligible.
To maximize the robustness of the system, particular emphasis must be laid on handling the \emph{aerodynamic influence} of the MAV itself, and the wind gusts.
First, the control subsystem must be resistant to these aerodynamic disturbances, in order to provide control stability (we rely on low-level stabilization designed in our team for flying in demanding desert conditions \cite{MBZIRC_magnetic_grasping}).
Second, perception modules must maintain their sensory properties when flying in low lighting conditions and in dust clouds, which originate when previously settled dust starts whirling.\looseness=-1

An obvious constraint arises from the particular \mbox{\emph{historical}} \mbox{\emph{value}} of the surveyed objects and their invaluable character, which is the key reason for undertaking the documentation and restoration work.
To avoid potential damage to fragile artifacts or to their surroundings at all costs, the MAV has to maximize the reliability and robustness attributes of the hardware and the software systems.
The introduction of redundancy, system fault detectors, and safety procedures is a critical requirement for such a \mbox{safety-challenging environment}.

Last but not least, the system has to provide high \emph{payload modularity}, in order to tackle all the documentation tasks in various environments.
During documentation works, it may be necessary to document vertical walls and also the ceiling, even with multiple types of payload.
For this reason, the hardware platform must be capable of rapidly changing the payload, its position, and the stabilization axes, in order to provide an effective solution to these end-user requirements.



\vspace*{-0.55em}
\section{Aerial Platform}
\begin{figure}[htb]
  \hspace*{-0.4em}
  \centering
  \subfloat{
    \includegraphics[width=0.49\columnwidth]{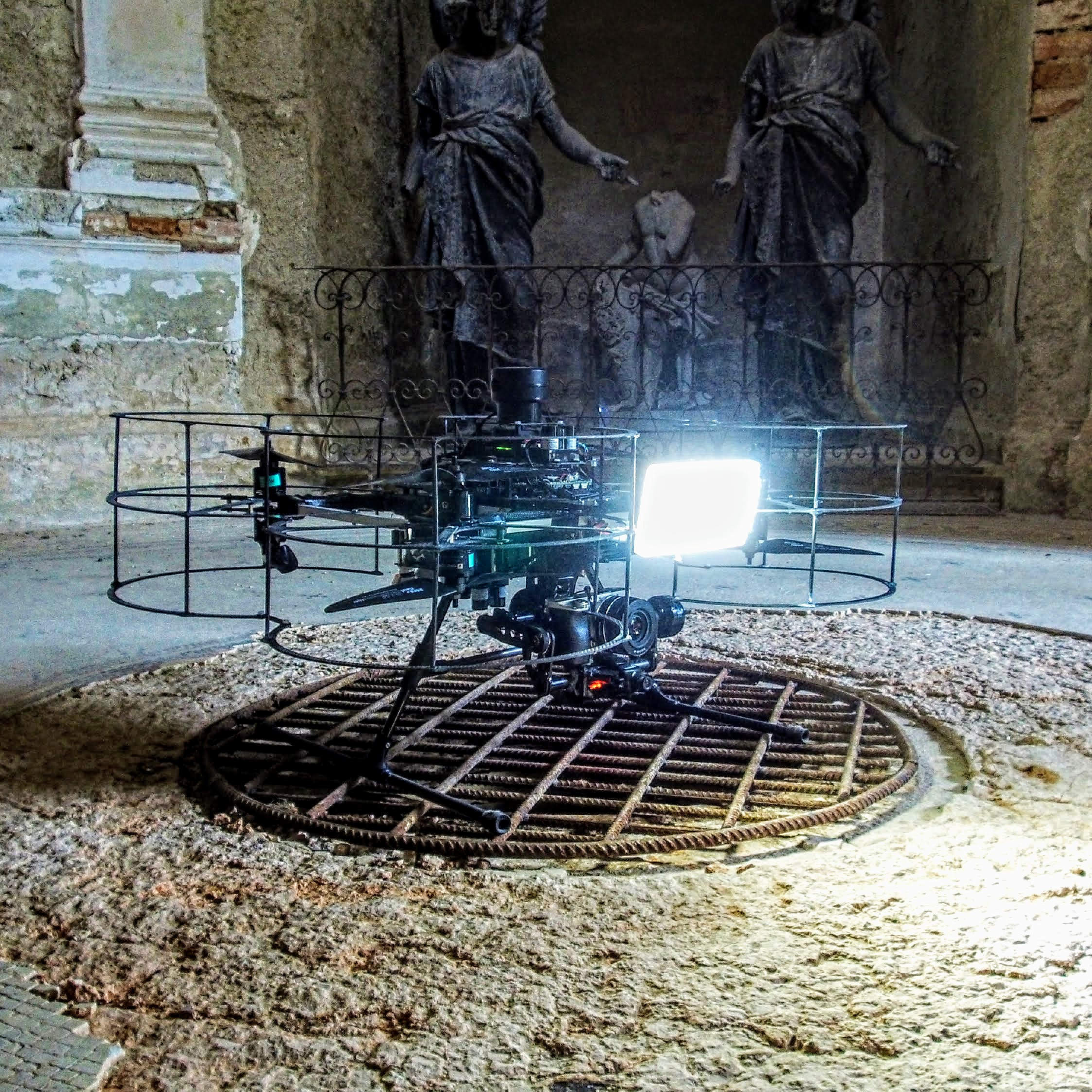}
  }
  \hspace*{-0.8em}
  \subfloat{
    \includegraphics[width=0.49\columnwidth]{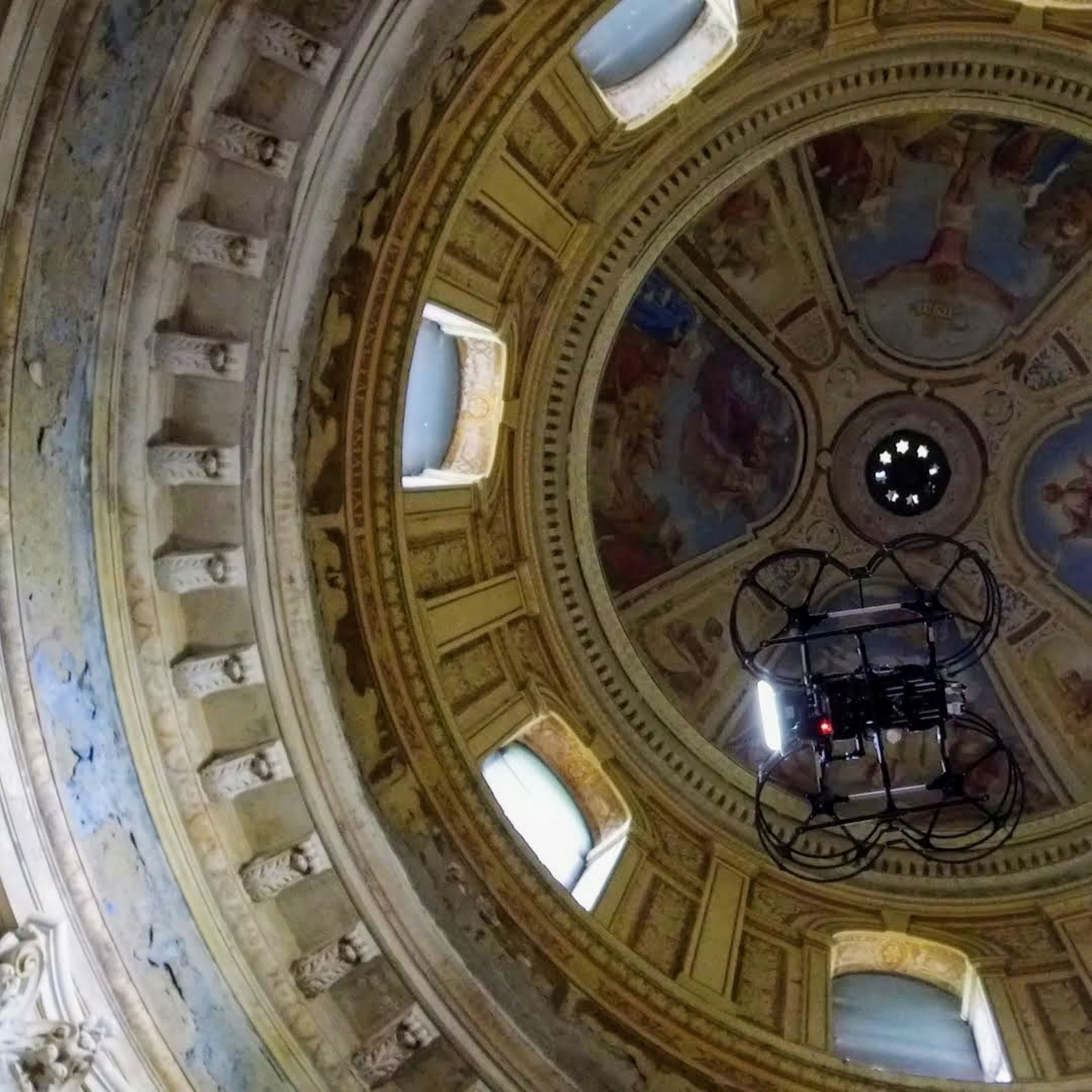}
  }
  \caption{An application-tailored MAV carrying sensory and mission equipment during documentation of the Klein mausoleum in Sobot\'{i}n, Czech Republic}
  \label{fig:aerial_platform}
\end{figure}

A specialized multi-MAV platform was developed to survey dark areas of historical monuments.
This system is distributed to the primary MAV carrying the mission payload and a couple of lighter MAVs carrying additional mission equipment, such as lights.
The primary MAV, shown in \autoref{fig:aerial_platform}, is designed to minimize its dimensions, since the task assumes flights in narrow passages close to obstacles.
Simultaneously, it is designed to maximize its payload weight capacities, since the payload is defined by the end-users, is interchangeable, and often cannot be optimized for employment on aerial platforms.
In its default configuration, the primary MAV carries an autopilot, an onboard computer, a down-oriented camera, two laser rangefinders, a 2-D laser scanner, a \threeD camera for obstacle detection, and the payload (an onboard light and a 2-axes stabilization hinge with a professional camera with its lens and a first-person view (FPV) system).
The total weight is \SI{5}{\kilogram}, with a payload weight of \SI{1.5}{\kilogram}.
To provide an extra level of safety, the platform is equipped with a mechanical propeller guard system to isolate the propellers from the external environment.
The lighter MAVs carry a light with a digitally-controlled pitch angle (a degree-of-freedom required for formation trajectory optimization in \cite{saska17etfa}), light intensity, and color warmth.

\vspace*{-1.5em}



\vspace*{-1.5em}
\section{System Architecture}
\label{sec:system_architecture}

\begin{figure*}
  \centering
  \usetikzlibrary{shapes.geometric,backgrounds,calc,arrows}
\usetikzlibrary{shadows}
\pgfdeclarelayer{background}
\pgfdeclarelayer{foreground}
\pgfsetlayers{background,main,foreground}

\tikzset{radiation/.style={{decorate,decoration={expanding waves,angle=90,segment length=4pt}}}}
\tikzstyle{block}=[draw, rounded corners, text centered, minimum height=1.4em, fill opacity=0.5, text opacity=1.0]
\def\nodedst{1.2cm}

\begin{tikzpicture}[auto, node distance=3.4cm, >=latex]

  \node [block] (navigation) {\footnotesize Mission Navigation};
  \node [block, right of=navigation, shift = {(0.1, -0.4)}] (mpc) {\footnotesize MPC Tracker \& Controller};
  \node [block, right of=mpc, node distance=4.1cm] (so3) {\footnotesize Acceleration Controller};
  \node [block, right of=so3, fill=white, node distance=4.1cm] (attitude) {\footnotesize Attitude Rate Controller};
  \node [block, right of=attitude, node distance=2.7cm] (actuator) {\footnotesize Actuators};
  \node [block, below of=so3, node distance=\nodedst] (state) {\footnotesize State Estimation};
  \node [block, below of=attitude, node distance=\nodedst] (data_preprocessing) {\footnotesize Data Preprocessing};
  \node [rectangle, below of=actuator, node distance=\nodedst, text width=5.5em, text centered] (sensory_data) {\footnotesize Onboard Sensory Data};
  \node [block, below of=mpc, node distance=\nodedst] (map) {\footnotesize Global Map};

  \node [block, below of=navigation, shift = {(1.2, -0.3)}, node distance=\nodedst, text width=4.4em] (sfd) {\footnotesize System Fault Detection};
  \node [rectangle, below of=navigation, shift = {(-1.2, -0.3)}, node distance=\nodedst, text width=4.8em, text centered, minimum height=1.4em, fill opacity=0.5, text opacity=1.0] (operation) {\footnotesize Mission Specifications};

  \node[inner sep=0,minimum size=0,below of=navigation, node distance=2.4em] (k) {}; 
  \draw [->] (navigation.-20) |- node[below, shift={(0.5, 0.05)}] {\footnotesize $\mathbf{r}_{d}, \psi_{d}$} (mpc.west);

  \draw [->] (mpc.east) -- node[below, shift={(0, 0.07)}] {\footnotesize $\ddot{\mathbf{r}}_d, \ddot{\psi}_d$}(so3.west);
  \draw [->] (so3.east) -- node[below, shift={(-0.1, 0)}] {\footnotesize $\boldsymbol{\omega}_d$, $T_d$} (attitude.west);
  \draw [->] (attitude.east)+(0.2,0) -- (actuator.west);
  \draw [->] (sfd.west) -| (navigation.south);

  \node[inner sep=0,minimum size=0,above of=operation, node distance=1.78em] (k) {}; 
  \draw [-, dashed] (operation.north) -- (k);
  \draw [->, dashed] (k) -| (navigation.220);

  \draw [->, dashed] (sensory_data.west) -- (data_preprocessing.east);
  \draw [->] (data_preprocessing.185) -- (state.-5);
  \draw [<->] (map.east) -- (state.west);
  \draw [->] (state.north) -- (so3.south);
  \draw [->] (state.north)+(0,0.47) node[right, shift={(-0.05, -0.25)}] {\footnotesize $\mathbf{x}, \boldsymbol{\omega}$} -| (mpc.south);
  \draw [->] (state.north)+(0,0.47) -| (navigation.-40);
  \draw [->] (attitude.south)+(0,-0.15) |- node[above, shift={(-1.2, -0.1)}] {\footnotesize $\mathbf{v}, \boldsymbol{\omega}, \mathbf{R}(\psi, \theta, \phi)$} +(-2.4, -0.61) |- (state.5);

  \begin{pgfonlayer}{background}
    \path (mpc.west |- mpc.north)+(-0.2,0.4) node (a) {};
    \path (actuator.south -| actuator.east)+(+0.2,-0.15) node (b) {};
    \path[fill=gray!1,rounded corners, draw=black!70, densely dotted]
      (a) rectangle (b);
  \end{pgfonlayer}
  \node [rectangle, above of=mpc, node distance=1.3em, shift={(-0.28,0.0)}] (text_control) {\footnotesize \textbf{Stabilization \& Control}};

  \begin{pgfonlayer}{background}
    \path (attitude.west |- attitude.north)+(-0.2,0.4) node (a) {};
    \path (attitude.south -| attitude.east)+(+0.2,-0.15) node (b) {};
    \path[fill=gray!3,rounded corners, draw=black!70, densely dotted]
      (a) rectangle (b);
  \end{pgfonlayer}
  \node [rectangle, above of=attitude, shift={(-0.0,0)}, node distance=1.2em] (autopilot) {\footnotesize \textbf{Autopilot}};

  \begin{pgfonlayer}{background}
    \path (map.west |- map.north)+(-0.2,0.4) node (a) {};
    \path (data_preprocessing.south -| data_preprocessing.east)+(+0.2,-0.1) node (b) {};
    \path[fill=gray!1,rounded corners, draw=black!70, densely dotted]
    (a) rectangle (b);
  \end{pgfonlayer}
  \node [rectangle, above of=map, node distance=1.3em, shift = {(-0.19, 0.0)}] (text_localization) {\footnotesize \textbf{Localization}};

\end{tikzpicture}
  \caption[Control Architecture]{
    High-level system pipeline of a single MAV.
    The stabilization \& control pipeline \cite{baca2018mpc} takes reference trajectory $\mathbf{r}_d, \psi_d$ (points sequence of the desired \threeD position and yaw) as a setpoint for the MPC in the MPC tracker, which outputs a command $\ddot{\mathbf{r}}_d,\;\ddot{\psi}_d$ for the acceleration tracking SO(3) controller.
    The acceleration controller produces the desired angular rate $\boldsymbol{\omega}_d$ and thrust reference $T_d$ for the embedded attitude rate controller.
    The localization pipeline is described in detail in \autoref{sec:state_estimation}.
  }
  \label{fig:system_architecture}
\end{figure*}
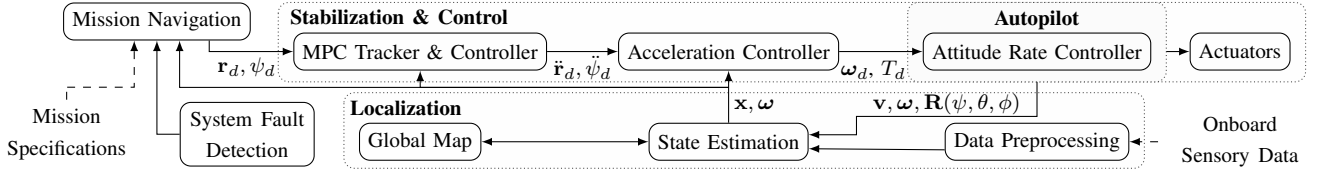

The overall system architecture is composed of four main subsystems, which are hereafter described.
The high-level pipeline of the system is outlined in {\autoref{fig:system_architecture}}.

\vspace*{-0.54em}
\subsection{Control Architecture}
\label{ssec:control_architecture}

An MAV disturbance-resistant control pipeline was developed in the previous work of our group \mbox{\cite{baca2016embedded, baca2018mpc, ecmr17_mbzirc_landing, MBZIRC_magnetic_grasping}}.
Beyond others, the MPC-based approach {\cite{baca2018mpc}} was tested in the harsh environment of the desert in the United Arab Emirates during the MBZIRC 2017 competition, where it outperformed 147 registered teams \mbox{\cite{ecmr17_mbzirc_landing, MBZIRC_magnetic_grasping}}.
The system architecture presented in {\autoref{fig:system_architecture}} is based on experience gained during this competition, which posed similar requirements of reliability, and resistance to wind disturbance and omnipresent dust.
However, the task presented here is considerably different due to absence of GNSS and the density of the obstacles, and therefore goes beyond the work presented in \mbox{{\cite{MBZIRC_magnetic_grasping}} and {\cite{ecmr17_mbzirc_landing}}}.




\vspace*{-0.54em}
\subsection{Localization}
\label{sec:state_estimation}

For 6 degrees-of-freedom state estimation, we propose to rely on three laser-based sensors.
First, a rigidly-mounted lightweight 2-D scanner produces data in the horizontal plane of the vehicle.
Second, two point-distance laser sensors (rangefinders) measure the distance to the ground and ceiling objects.
Together with an onboard IMU and a sparse \mbox{\threeD} map, the laser-inertial approach manages to estimate the global position and the attitude in light-independent conditions and without any heavyweight sensory equipment.
The whole localization pipeline is summarized in \autoref{fig:state_estimation}, and will be described in detail in this section.

\subsubsection{Global Map}
\label{ssec:global_map}

The localization system proposed in this work is designed to operate with a partially-known map registered from multiple \mbox{3-D} intensity/color point clouds produced by a terrestrial laser scanner (TLS).
The necessary granularity of the map ({\SI{10}{\centi\metre}} in our experiments) depends on the onboard computational resources and on the structural complexity of the documented building.
Nonetheless, a modern TLS is capable of producing a scan with millimeter-level granularity, which makes the map subject to data reduction.

A raw map is processed by a set of filters (median filter, outlier rejection, uniform sampling) and is transformed to octree representation in order to employ optimized map operations, such as node traversal, integration of sensor measurements, data access, and tree node queries.
During the preprocessing phase, artificial ground data is injected into the map to cope with missing data due to occlusions during scanning.
Assuming that the ground is a cavity-free plane, the ground data is augmented by a set of points uniformly sampled from a plane.
The parameters of this sampled plane are obtained from fitting it on a set of points withdrawn from the undermost parts of the available map using the RANSAC algorithm.
Missing non-ground data is augmented in midair during a mission by the map refinement module.

A smaller structure requires \mbox{\textless$\,$5} scans, where a single full-dome scan ({\SI{360}{\degree}} horizontal and {\SI{300}{\degree}} vertical field of view) takes approximately {\SI{3}{\minute}}.
During a field operation, there is enough time to produce a map of the object during the preparation of the equipment required for the mission.
Example of a map is shown on the right side of \mbox{{\autoref{fig:motivation_title}} and in {\autoref{fig:chlumin_octree}}}.


\vspace*{0.05em}
\subsubsection{State Estimation}
An MAV is assumed to have first-order dynamics for a short period of time during hovering and slow flights with negligible tilts (these flight characteristics are required in the confined areas in historical monuments for safety reasons).
The linear stochastic discrete state-space model is used as
\vspace*{-0.2em}
\begin{align}
  \mathbf{x}_{[k]} &= \mathbf{A}_{[k]}\mathbf{x}_{[k-1]} + \mathbf{B}_{[k]}\mathbf{u}_{[k]} + \mathbf{\eta}_{[k]},\\
  \mathbf{z}_{[k]} &= \mathbf{H}_{[k]}\mathbf{x}_{[k]} + \mathbf{\upsilon}_{[k]}.
\end{align}
The state $\mathbf{x}_{[k]}$, system input $\mathbf{u}_{[k]}$, measurements $\mathbf{z}_{[k]}$ and random noises of the system at time $k$ are given as
\begin{align}
  \mathbf{x}_{[k]} &= \left(\mathbf{p}_{[k]}^T,\;\mathbf{\Omega}_{[k]}^T\right)^{T},\;  \mathbf{u}_{[k]} = \left(\mathbf{v}_{[k]}^{T},\;\boldsymbol{\omega}_{[k]}^{T}\right)^T,\\
  \mathbf{z}_{[k]} &= \hat{\mathbf{x}}_{[k]},\;\;\;\mathbf{\eta}_{[k]} \sim \mathcal{N}(0,\,\mathbf{Q}_{[k]}),\;\;\;\mathbf{\upsilon}_{[k]} \sim \mathcal{N}(0,\,\mathbf{R}_{[k]}),
\end{align}
where the state $\mathbf{x}_{[k]}$ is comprised from global position $\mathbf{p}_{[k]} = \left(x_{[k]},\;y_{[k]},\;z_{[k]}\right)^T$ and attitude $\mathbf{\Omega}_{[k]} = \left(
\psi_{[k]},\;\theta_{[k]},\;\phi_{[k]}\right)^T$ consisting of the yaw, pitch, and roll angles; $\mathbf{v}_{[k]}$ is the linear and $\boldsymbol{\omega}_{[k]}$ is the angular velocity of the IMU frame; $\hat{\mathbf{x}}_{[k]}$ is the measured global state; and $\mathbf{S}_{[k]}$ and $\mathbf{Q}_{[k]}$ are the covariance matrices of the process and the measurement noise at time $k$.
The state-transition model $\mathbf{A}_{[k]}$, the control-input model $\mathbf{B}_{[k]}$, the observation model $\mathbf{H}_{[k]}$, and the covariance matrices $\mathbf{S}_{[k]}$ and $\mathbf{Q}_{[k]}$ are defined as
\begin{align}
  \mathbf{A}_{[k]} &= \begin{bmatrix}
    \mathbf{I}_{6 \times 6}
    \end{bmatrix},\mathbf{Q}_{[k]} = \Delta t_{[k]}\begin{bmatrix}
    \mathbf{\Sigma}^{mcl}_{6 \times 6} & \mathbf{0}_{6 \times 6}\\
    \mathbf{0}_{6 \times 6} & \mathbf{\Sigma}_{6 \times 6}^{icp}
  \end{bmatrix},\\
    \mathbf{B}_{[k]} &= \Delta t_{[k]}\begin{bmatrix}
      \mathbf{R}(\psi_{[k]}, \theta_{[k]}, \phi_{[k]})_{3 \times 3} & \mathbf{0}_{3 \times 3}\\
      \mathbf{0}_{3 \times 3} & \mathbf{R}(\psi_{[k]}, \theta_{[k]}, \phi_{[k]})_{3 \times 3}
    \end{bmatrix},\nonumber\\
      \mathbf{S}_{[k]} &= \Delta t_{[k]}\begin{bmatrix}
        \mathbf{\sigma}_{\mathbf{p}}^2\mathbf{I}_{3 \times 3} & \mathbf{0}_{3 \times 3}\\
        \mathbf{0}_{3 \times 3} & \mathbf{\sigma}_{\mathbf{\Omega}}^2\mathbf{I}_{3 \times 3}
      \end{bmatrix},\;
      \mathbf{H}_{[k]} = \begin{bmatrix}
        \mathbf{I}_{6 \times 6} & \mathbf{I}_{6 \times 6}
      \end{bmatrix}^T,\nonumber
      \end{align}
      where $\mathbf{I}_{n \times n} \in \mathbb{R}^{n \times n}$ is an identity matrix and \mbox{$\mathbf{0}_{n \times n} \in \mathbb{R}^{n \times n}$} is an empty matrix, $\boldsymbol{\Sigma^{\bullet}}_{6 \times 6} \in \mathbb{R}^{6 \times 6}$ is the covariance matrix of the global and local state estimation, \mbox{$\Delta t_{[k]} = t_{[k]} - t_{[k-1]}$} is the time elapsed since the last KF update, and $\mathbf{R}(\psi_{[k]}, \theta_{[k]}, \phi_{[k]}) \in \mathbb{R}^{3\times 3}$ is the \threeD attitude.
      The presence of the rotation matrix in the control-input model copes with the differing global and IMU frames.
      The input of the system consists of inertial measurements coming at \SI{100}{\Hz}, and observations are produced by two estimation processes running in parallel, incoming at \SIlist[list-units=single,list-final-separator = {and }, list-pair-separator= { and }]
      {5;20}{\Hz}, which will be described below.
      The output of the KF correction step is equal to the output of the whole localization process.


      \begin{figure*}
        \centering
        \usetikzlibrary{shapes.geometric,backgrounds,calc,arrows}
\usetikzlibrary{shadows}
\pgfdeclarelayer{background}
\pgfdeclarelayer{foreground}
\pgfsetlayers{background,main,foreground}

\tikzset{radiation/.style={{decorate,decoration={expanding waves,angle=90,segment length=4pt}}}}
\tikzstyle{block}=[draw, text width=3.8em, rounded corners, text centered, minimum height=1.2em, fill opacity=0.5, text opacity=1.0]
\def\nodedst{0.95cm}

\hspace*{-0.5em}
\begin{tikzpicture}[auto, node distance=2.1cm, >=latex]

  \node [block] (outlier_removal) {\footnotesize Outlier Removal};
  \node [block, right of=outlier_removal] (voxel_sampling) {\footnotesize Voxel Sampling};
  \node [block, below of=voxel_sampling, node distance=1.0cm, shift={(-1.05, 0)}] (median_filter_down) {\footnotesize Median\\Filter};
  \node [block, below of=median_filter_down, node distance=1.0cm] (median_filter_up) {\footnotesize Median\\Filter};

  \node [block, left of=outlier_removal, node distance=2.4cm] (2d_sensor) {\footnotesize 2-D Laser Scanner};
  \node [block, below of=2d_sensor, node distance=1.0cm] (rangefinder_down) {\footnotesize Rangefinder down};
  \node [block, below of=rangefinder_down, node distance=1.0cm] (rangefinder_up) {\footnotesize Rangefinder up};

  \node [block, right of=voxel_sampling, node distance=2.5cm, shift={(0.55em, 1.7em)}, text width=5.5em] (map) {\footnotesize Map Refinement};

  \node [block, below of=map, node distance=\nodedst*4/3, shift={(-0.55em, 0)}] (adaptive_sampling) {\footnotesize Adaptive Sampling};
  \node [block, right of=adaptive_sampling] (motion_model) {\footnotesize Motion Model};
  \node [block, right of=motion_model, shift = {(-0.05, 0.0)}] (observation_model) {\footnotesize Observation Model};
  \node [block, right of=observation_model, shift = {(0.05, 0.0)}] (mcl_state_estimation) {\footnotesize State Estimation};
  \node [block, below of=adaptive_sampling, node distance=\nodedst*7/6] (dead_reckoning) {\footnotesize Dead Reckoning};

  \node [block, right of=dead_reckoning, shift={(0, -\nodedst*5/12)}, text width=4.9em] (scan_bundle) {\footnotesize Scan Bundle};
  \node [block, right of=scan_bundle, shift={(-0.05,\nodedst/2)}] (vertical_estimation) {\footnotesize Vertical};
  \node [block, below of=vertical_estimation, node distance=\nodedst*2/4] (lateral_estimation) {\footnotesize Lateral};
  \node [block, right of=vertical_estimation, shift={(0.05,-\nodedst/4)}] (icp_state_estimation) {\footnotesize State Estimation};

  \node [block, right of=mcl_state_estimation, node distance=2.5cm, text width=4.9em, shift={(0, 0.44)}] (lkf) {\footnotesize Linear Kalman Filter};

  \node [block, below of=lkf, node distance=0.97cm] (ekf) {\footnotesize EKF};
  \node [block, below of=ekf, node distance=\nodedst*2/3] (imu) {\footnotesize IMU 1-3};


  \draw [->] (2d_sensor.east) -- node[above]{\scriptsize \SI{20}{\Hz}} (outlier_removal.west);
  \draw [->] (rangefinder_down.east) -- node[above]{\scriptsize \SI{100}{\Hz}} (median_filter_down.west);
  \draw [->] (rangefinder_up.east) -- node[above]{\scriptsize \SI{100}{\Hz}} (median_filter_up.west);
  \draw [->] (outlier_removal.east) -- (voxel_sampling.west);
  \draw [->] (adaptive_sampling.east) -- (motion_model.west);
  \draw [->] (motion_model.east) -- (observation_model.west);
  \draw [->] (observation_model.east) -- (mcl_state_estimation.west);
  \draw [->] (scan_bundle.east) |- (lateral_estimation.west);

  \node[inner sep=0,minimum size=0,right of=mcl_state_estimation, node distance=2.8em] (k) {}; 
  \draw [-] (mcl_state_estimation.east) -- (k);
  \draw [->] (k) |- node[above, shift={(0.25, -0.05)}] {\scriptsize $\mathbf{x}^{mcl}$} (lkf.165);

  \node[inner sep=0,minimum size=0,right of=icp_state_estimation, node distance=3.5em] (k) {}; 
  \draw [-] (icp_state_estimation.east) -- (k);
  \draw [->] (k) |- node[above, shift={(0.04, -0.08)}] {\scriptsize $\mathbf{x}^{icp}$} (lkf.195);

  \draw [->] (imu.north) -- (ekf.south);

  \node[inner sep=0,minimum size=0,right of=median_filter_down, node distance=6.7em] (k) {}; 
  \node[inner sep=0,minimum size=0,right of=median_filter_down, node distance=5.6em] (kk) {}; 
  \node[inner sep=0,minimum size=0,left of=adaptive_sampling, node distance=2.5em] (y_mcl) {}; 
  \node[inner sep=0,minimum size=0,left of=scan_bundle, node distance=3em, shift={(0, -0.6em)}] (y_icp) {}; 
  \node[inner sep=0,minimum size=0,below of=k, node distance=3.93em] (y_icp_left) {}; 
  \draw [-] (median_filter_down.east) -- (kk);
  \draw [-] (median_filter_up.east) -| (kk);
  \draw [-] (voxel_sampling.east) -| (kk);
  \draw [-, double] (kk) node[right, shift = {(0.35, -0.00)}]{\scriptsize $\mathbf{y}$} -| (k);
  \draw [-, double] (k) |- (map.west);
  \draw [->, double] (k) |- (y_mcl);
  \draw [-, double] (k) -- (y_icp_left);
  \draw [->, double] (y_icp_left) -- (y_icp);

  \draw [->] (lkf.north) |- (map.east);
  \draw [->] (ekf.north) -- node[left] {\scriptsize $\mathbf{v}, \boldsymbol{\omega}$} node[right, shift = {(0.0, 0.02)}] {\scriptsize $\mathbf{R}$} (lkf.south);

  \node[inner sep=0,minimum size=0,left of=ekf, node distance=3.0em] (j) {}; 
  \node[inner sep=0,minimum size=0,left of=dead_reckoning, node distance=2.9em] (l) {}; 
  \node[inner sep=0,minimum size=0,below of=l, node distance=2.15em] (k) {}; 
  \node (m) at (intersection of k--l and y_icp_left--y_icp) {};
  \draw [-] (ekf.west) -- (j);
  \draw [-] (j) |- (k);
  \draw [-] (k) -- (m);
  \draw [-] (m) -- (l);
  \draw [->] (l) -- (dead_reckoning.west);

  \node[inner sep=0,minimum size=0,right of=dead_reckoning, node distance=2.6em] (j) {}; 
  \node[inner sep=0,minimum size=0,left of=vertical_estimation, node distance=3.2em] (k) {}; 
  \draw [-] (dead_reckoning.east) -- (j);
  \draw [->] (j) |- (motion_model.195);
  \draw [-] (j) -| (k);
  \draw [->] (k) -- (vertical_estimation.west);
  \draw [->] (j) |- (scan_bundle.west);

  \node[inner sep=0,minimum size=0,left of=icp_state_estimation, node distance=3.0em] (k) {}; 
  \draw [-] (lateral_estimation.east) -| node[below, shift={(-0.03, 0.1)}]{\scriptsize $y, \psi$} node[above, shift={(-0.15, -0.08)}]{\scriptsize $x,$} (k);
  \draw [-] (vertical_estimation.east) -| node[above, shift={(-0.1, -0.05)}]{\scriptsize $z$} (k);
  \draw [->] (k) -- (icp_state_estimation.west);

  \draw [<-] (mcl_state_estimation.220) -- node[left, shift = {(0.08, -0.11)}]{\scriptsize $\mathbf{\epsilon}_{icp}$} (icp_state_estimation.140);
  \draw [->] (mcl_state_estimation.-40) -- node[left, shift = {(0.08, -0.08)}]{\scriptsize $\mathbf{\Sigma}^{mcl}$} node[right, shift = {(-0.04, -0.08)}]{\scriptsize $\mathbf{x}^{mcl}_{0}$} (icp_state_estimation.40);

  \node[inner sep=0,minimum size=0,above of=lkf, node distance=2.35em, shift={(0.5, 0)}] (output) {}; 
  \draw [->] (lkf.north) |- node[above, shift={(0.2, -0.05)}] {\scriptsize $\mathbf{x}$} (output);


  \begin{pgfonlayer}{background}
    \path (2d_sensor.west |- 2d_sensor.north)+(-0.1,0.4) node (a) {};
    \path (median_filter_up.south -| voxel_sampling.east)+(+0.3,-0.1) node (b) {};
    \path[fill=gray!3,rounded corners, draw=black!70, densely dotted]
    (a) rectangle (b);
  \end{pgfonlayer}
  \node [rectangle, above of=outlier_removal, shift={(-2.4em,0)}, node distance=1.8em] (onboard sensors) {\footnotesize \textbf{Onboard Sensors Data Pre-processing}};

  \begin{pgfonlayer}{background}
    \path (ekf.west |- ekf.north)+(-0.1,0.05) node (a) {};
    \path (imu.south -| imu.east)+(+0.1,-0.4) node (b) {};
    \path[fill=gray!3,rounded corners, draw=black!70, densely dotted]
    (a) rectangle (b);
  \end{pgfonlayer}
  \node [rectangle, below of=imu, shift={(-0.0em,0)}, node distance=1.25em] (autopilot) {\footnotesize \textbf{Autopilot}};

  \begin{pgfonlayer}{background}
    \path (adaptive_sampling.west |- adaptive_sampling.north)+(-0.1,0.4) node (a) {};
    \path (mcl_state_estimation.south -| mcl_state_estimation.east)+(+0.1,-0.1) node (b) {};
    \path[fill=gray!3,rounded corners, draw=black!70, densely dotted]
    (a) rectangle (b);
  \end{pgfonlayer}
  \node [rectangle, above of=motion_model, shift={(-3.7em,0)}, node distance=1.9em] (mcl) {\footnotesize \textbf{Monte Carlo Localization}};

  \begin{pgfonlayer}{background}
    \path (scan_bundle.west |- icp_state_estimation.north)+(-0.1,0.2) node (a) {};
    \path (icp_state_estimation.south -| icp_state_estimation.east)+(+0.1,-0.1) node (b) {};
    \path[fill=gray!3,rounded corners, draw=black!70, densely dotted]
    (a) rectangle (b);
  \end{pgfonlayer}
  \node [rectangle, above of=lateral_estimation, shift={(-6.0em,0)}, node distance=1.85em] (scan_matching) {\footnotesize \textbf{Scan Matching}};
\end{tikzpicture}
        \caption[State Estimation]{
          Workflow diagram of the state estimation process.
          The laser-inertial pipeline fuses a global map, onboard data from multiple laser-based sensors (a 2-D horizontal scanner and an up- and down-oriented rangefinder), and 3 IMUs to produce state estimation $\mathbf{x}$.
          The IMUs are fused by an Extended Kalman Filter outputting linear $\mathbf{v}$ and $\boldsymbol{\omega}$ angular velocity, and attitude $\mathbf{R}(\psi, \theta, \phi)$, where $\psi, \theta, \phi$ are the Euler angles yaw, pitch, and roll.
          A decoupled local state refinement employing Iterative Closest Point is initialized after a global state estimate $\mathbf{x}_{0}^{\;mcl}$ is provided by MCL.
          The validity of the state estimation processes is observed with respect to the MCL estimate covariance $\boldsymbol{\Sigma}^{mcl}$ and the absolute mean square error $\epsilon_{icp}$ of the scan matching.
        }
        \label{fig:state_estimation}
      \end{figure*}
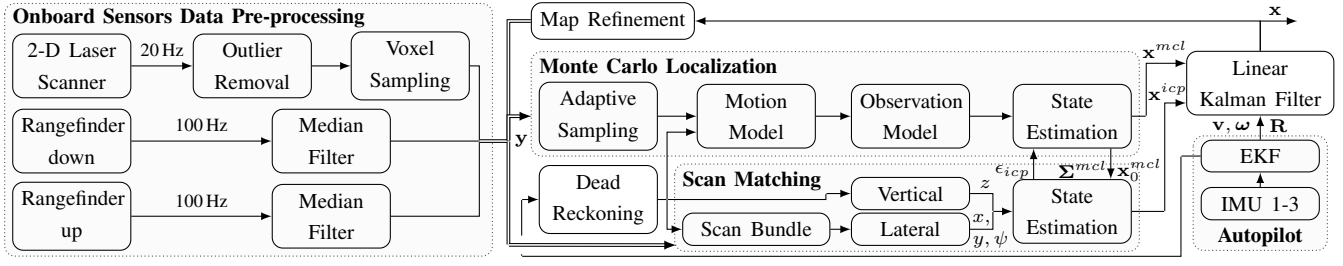

      \vspace*{0.05em}
      \subsubsection{Monte Carlo Localization}
      The configuration space of a robot inside an a-priori known map of a historical object is immense.
      This restricts the straight registration of sensory data to the extensive map due to the unknown initial conditions, which MCL provides in the form of a slow global state estimate.
      Concisely, MCL determines the posterior probability $p(\mathbf{x}|\mathbf{y}, \mathbf{u})$ of an unobservable state $\mathbf{x}$ given sensor observations $\mathbf{y}$ and control inputs $\mathbf{u}$ by computing it on the state space subset in the form of hypotheses, yielding an approximation of the probability density function.
      The posterior probability can be obtained by employing the Bayes filter, which recursively computes the previous equation in the form of a belief $Bel(\mathbf{x})$ of the posterior \mbox{probability as}
      \begin{align}
        Bel(\mathbf{x}) = \eta\; p(\mathbf{y}|\mathbf{x})\int p(\mathbf{x}|\mathbf{\hat{x}}, \mathbf{u}) Bel(\mathbf{\hat{x}}) d\mathbf{\hat{x}},
      \end{align}
      where $\eta$ is a normalization constant.
      The derivation of the equation holds under the initial condition $p(\mathbf{x}_{0}) = p(\mathbf{x}_{0}|\mathbf{y}_{0}, \mathbf{u}_{0})$ and Markov independence assumptions.


      \vspace*{0.05em}
      \textit{Motion model:}
      An odometry-based model for 2-D mobile robots employing the dead-reckoning principle is expanded to 3-D.
      In comparison with \cite{motion_model_3d}, our application requires slow movement of an airborne vehicle up to \SI{0.5}{\metre\per\second}, making the variations in roll and pitch negligible and therefore allowing us to reduce the kinematic DoF to 4 (\threeD position and heading).

      \textit{Adaptive sampling:}
      To improve performance, KLD-sampling \cite{Fox:2001:KAP:2980539.2980632} estimates the sufficient number of hypotheses $M$ by bounding the error introduced by the sample-based representation of the MCL.
      The estimate is based on drawing from a discrete distribution with $p$ different bins, and for
      \begin{align}
        M \approx \frac{p-1}{2\epsilon}\left(1+\frac{2}{9(p-1)} + \sqrt{\frac{2}{9(p-1)}}z_{1-\delta}\right)^3,
      \end{align}
      guarantees with probability $1-\delta$ that the Kullback--Leibler distance between the maximum likelihood estimate (MLE) and the true distribution is less than $\epsilon$, with $z_{1-\delta}$ being the upper $1-\delta$ quantile of the normal $\mathcal{N}(0, 1)$ distribution.

      To prevent convergence to an erroneous local minimum, a subset of hypotheses with the lowest weights is replaced in each resampling step with
      a dynamic-size set of new randomly generated hypotheses over the whole sampling space and
      a static-size set of new hypotheses matching the position of the latest state estimate with randomly sampled heading.
      The ratio of newly injected hypotheses is regulated by Augmented-MCL \cite{thrun2005probabilistic}, which compares the short-term and long-term likelihood of observations as
      \begin{align}
        M_{new} = M\max\left\{0, 1 - \frac{w_{fast}}{w_{slow}}\right\},
      \end{align}
      where $w_{slow} = w_{slow} + \alpha_{slow}(\overline{w} - w_{slow})$ and $w_{fast} = w_{fast} + \alpha_{fast}(\overline{w} - w_{fast})$ for $\overline{w}$ being the weighted average over the whole set of hypotheses, and $0 \le \alpha_{slow} \ll \alpha_{fast}$ are the decay rates.


      \subsubsection{Local Refinement}
      To obtain precise and fast localization, local map registration is performed in a decoupled manner.
      The decoupling emerges from the sensory setup due to the vast difference between the data volume in the horizontal and in the vertical plane.
      In contrast to the vertical plane, where only two point-distance measurements are obtained, the horizontal sensor generally provides a greater number of samples (e.g., 16K samples per second for RPLIDAR A3), which needs to be reduced.
      The vast difference in the data volumes requires decoupling, otherwise the horizontal estimation would heavily overweigh the vertical estimation.

      \textit{Lateral estimation} employs a variant of the Iterative Closest Point (ICP) algorithm.
      Given a reference set of points $\mathbf{P}$ and a target set of points $\mathbf{Q}$, the optimization process finds a transformation $\mathbf{T}$, which minimizes the weighted point-to-point error metric
      \begin{align}
        E(\mathbf{T}) = \frac{1}{N}\sum_{i=1}^{N}(w_i\,||\mathbf{T}\mathbf{p}_i - \mathbf{q}_i||^2)
        \label{eq:icp_cost_function}
      \end{align}
      over the set of $N$ correspondence pairs $(\mathbf{p}_i, \mathbf{q}_i, w_i),$ $\mathmbox{\mathbf{p}_i \in \mathbf{P}},\;\mathbf{q}_i \in \mathbf{Q},\;w_i \in \mathbb{R},\;\forall i \in \left<1,\;N\right>,\;N \in \mathbb{Z}$.
      An initial solution to \autoref{eq:icp_cost_function} is given by the dead-reckoning principle.
      Determining of the correspondence pairs involves closest distance pairing and a median filter, duplicate reference matches, and RANSAC-based pairs rejectors.
      Implementation of the ICP is based on the Point-Cloud library {\cite{pcl_registration}}.

      The reference scan $\mathbf{P}$ is obtained onboard from a \mbox{2-D} laser scanner, and its data are prepared according to \autoref{fig:state_estimation}.
      To provide improved robustness, a short history of the measurements is bundled together using short-time IMU-based dead-reckoning odometry, and is used as the reference scan $\mathbf{P}$ for \threeD scan matching.
      An example of the scan bundle, registered into a map in the form of a \threeD point cloud, is displayed on the right side of \autoref{fig:motivation_title}.
      The target scan $\mathbf{Q} = \mathbf{Q}_{pla} \smallsetminus \mathbf{Q}_{occ}$, $\mathbf{Q}_{pla} \in \mathbf{Q}_{map}$, $\mathbf{Q}_{occ} \in \mathbf{Q}_{pla}$, is derived from $\mathbf{Q}_{map}$ and state estimate from the previous time step $\mathbf{x}_{[k-1]}$.
      The subset $\mathbf{Q}_{pla}$ represents points of the map located in between two planes  parallel to the \textit{x-y} plane of the 2-D sensor frame at distance $\pm d_{pla}$ on the \textit{z} axis of the same frame.
      The subset $\mathbf{Q}_{occ} \in \mathbf{Q}_{pla}$ represents all visually occluded points for which the linear path of a laser beam from a sensor position (rigidly defined by $\mathbf{x}_{[k-1]}$) to $\mathbf{q} \in \mathbf{Q}_{occ}$ is collision-free.
      A ray-casting algorithm, implemented over an octree representation of the map, is employed to determine the collision status.

      During \textit{vertical estimation}, a lateral estimate of the \mbox{\textit{x, y}} axes, an attitude estimate, and the up- and down-oriented point-distance measurements are used to define a quadratic least squares problem
      \begin{align}
        z^* = \arg\min_{z \in \mathbb{R}}
        \big ( &\alpha_{\uparrow}(\hat{\mathbf{p}}(z),\;y_{\uparrow}^{r}) \;||y^{m}_{\uparrow}(\hat{\mathbf{p}}(z)) - y_{\uparrow}^{r}||^2 +\\+\;&\alpha_{\downarrow}(\hat{\mathbf{p}}(z),\;y_{\downarrow}^{r}) \;||y^{m}_{\downarrow}(\hat{\mathbf{p}}(z)) - y_{\downarrow}^{r}||^2 \big ) \nonumber
      \end{align}
      to find vertical \textit{z} axis position $z^*$, where $\hat{\mathbf{p}}(z) = \left(\textit{x}, \textit{y}, z\right)^T$, \mbox{$y^r_{\bullet}$ are} real sensor data, and $y^m_{\bullet}$ are map measurements found by map ray-casting.
      Bear in mind that the attitude and the rigid IMU-sensor transformations are neglected here to maintain simplicity.
      The validity coefficients $\alpha_{\bullet}$  are defined as\looseness=-1
      \begin{align}
        \alpha_{\bullet}(\hat{\mathbf{p}}(z),\;y_{\bullet}^r) =
        \begin{cases}
          0, & \text{if } y^m_{\bullet}(\hat{\mathbf{p}}(z)) \text{ or } y^r_{\bullet} \text{ is invalid},\\
          1, & \text{otherwise}.
        \end{cases}
      \end{align}

      \noindent
      Data invalidity emerges directly from invalid sensor measurements or from the absence of a map reference.
      In addition, the down-oriented sensor detects dynamic obstacles, such as people or map changes, which are observable from an identifiable discrepancy between real and map-based observations.
      These detections likewise classify the observations as invalid.
      In the case of $\alpha_{\uparrow} = \alpha_{\downarrow} = 0$, the $z$ axis prediction at time $k$ is given as
      \vspace*{-0.5em}
      \begin{align}
        z_{[k]} = z_{[k-1]} + z^{imu}_{[k]} - z^{imu}_{[k-1]},
      \end{align}
      where $z^{imu}$ represents the integrated \textit{z} axis position derived from the IMU-based dead-reckoning odometry.



      \vspace*{-0.3em}
      \subsection{Mission Navigation}
      \label{ssec:mission_navigation}
      To maximize robustness of the system, a visibility-constrained navigation is employed such that an MAV is allowed to maneuver only to obstacle-free areas visible from a front-facing depth camera.
      This approach supervises lidar-based perception by a redundant check for local obstacles in the camera field-of-view.
      An MPC-based control for navigation and trajectory optimization for MAV formations in the documentation task is introduced in our previous work {\cite{saska17etfa}}.

      \vspace*{-0.3em}
      \subsection{System Fault Detection}
      \label{ssec:system_fault_detection}
      In parallel to the mission controller, a tightly coupled fault detection system supervises all aspects of the mission.
      That includes supervision of the sensors and battery life status, of the state estimation covariance, or of the divergence from a preplanned trajectory.
      The whole system is implemented as a centralized high-level state machine capable of overriding the mission with an appropriate reaction to fault scenarios.
      Examples of these safety procedures are enforced controlled landing, trajectory execution termination, or manual take over of the control by a human operator.
      These safety responses can be likewise triggered by a mission operator, who is required to supervise the mission by an aviation authority.\looseness=-1


      \vspace*{-0.15em}
      \section{Experimental Evaluation}
      \label{sec:experimental_evaluation}
      To prove concept of the proposed method, the system was thoroughly verified in simulation (Gazebo 9 coupled with ROS Melodic), before it was deployed in position control feedback loop of an MAV.
      The main intention of the simulation was to estimate suitability of the developed system for deployment in safety-critical environments of historical buildings, to reduce probability of failures, and to obtain a qualitative analysis of the system behavior.
      Although the simulation results are omitted here due to lack of space, they can be found in \cite{mrs_dronument}.



      \vfill
      \pagebreak
      \subsection{Localization Precision Analysis}
\begin{figure}[H]
        \vspace*{-1.4em}
        \hspace*{-0.6em}
        \centering
        \subfloat[Reflector-mounted platform]{
          \includegraphics[width=0.499\columnwidth]{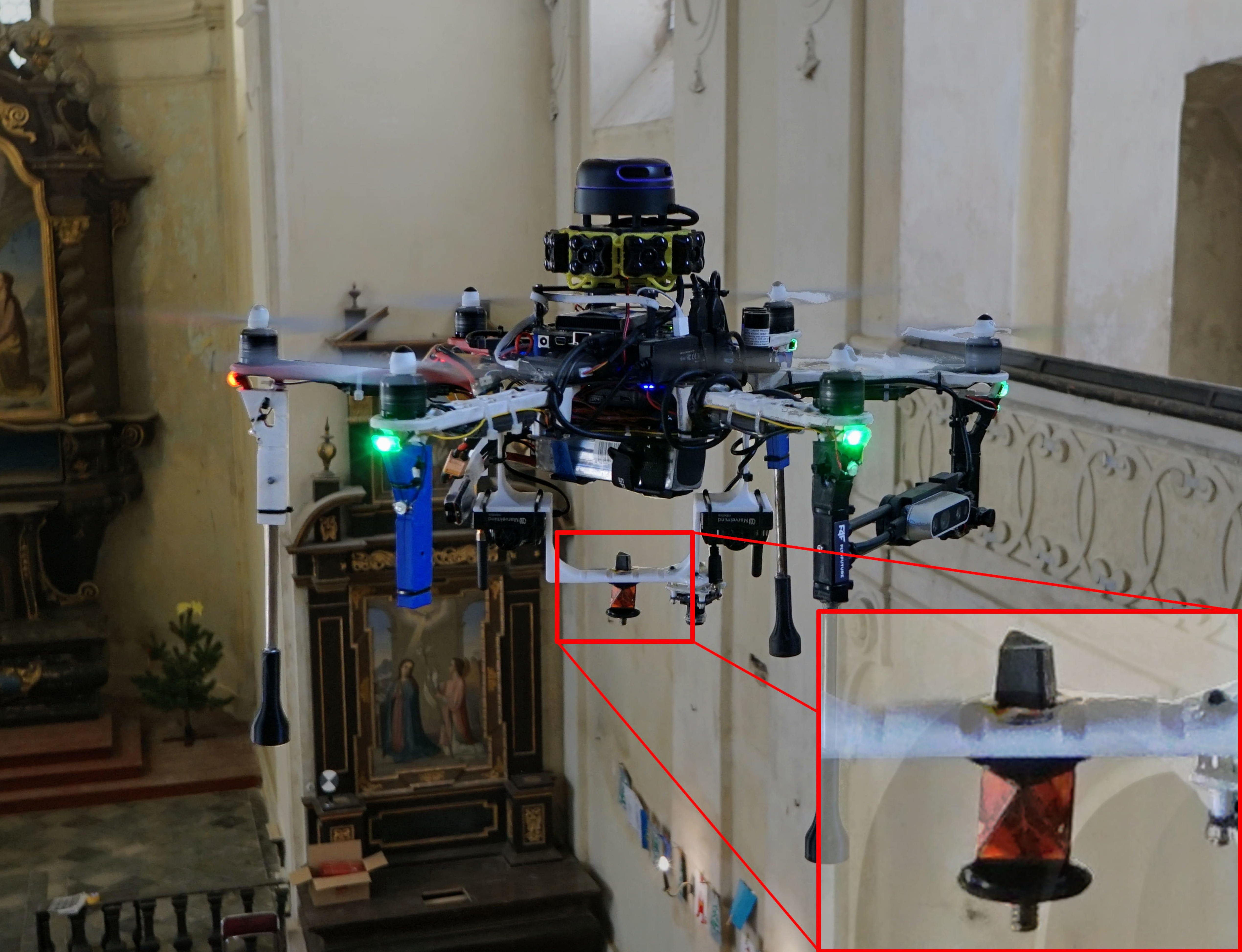}
          \label{fig:ground_truth_dataset_a}
        }
        \hspace*{-0.85em}
        \subfloat[Automatic tracking demonstration]{
          \includegraphics[width=0.499\columnwidth]{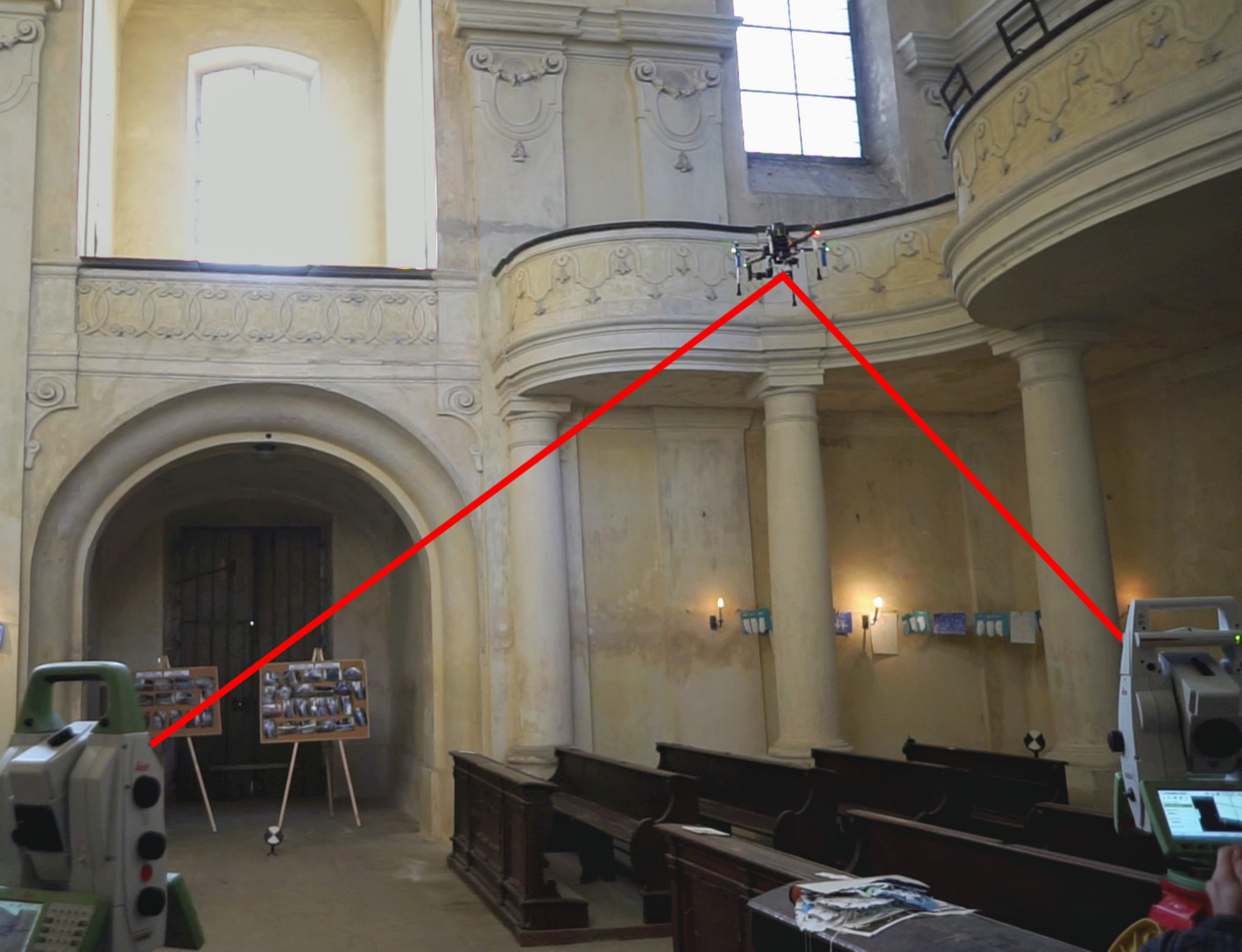}
          \label{fig:ground_truth_dataset_b}
        }
        \caption{An MAV platform equipped with onboard sensors and a reflector tracked by two Leica multi-stations measuring its \threeD position with precision of \SI{1.5}{\milli\meter} at \SI{5}{\Hz}}
        \label{fig:ground_truth_dataset}
        \vspace*{-0.8em}
    \end{figure}

      This section presents quantitative results of the localization system evaluated inside real church of St. Mary Magdalene in Chlum\'{i}n, using a prototype MAV with the same sensory setup as is carried by the presented project platform.
      To obtain ground truth data, two Leica multi-stations were employed to track movement of the MAV equipped with the Leica GRZ101 \SI{360}{\degree} Mini Prism reflector, as shown in \autoref{fig:ground_truth_dataset_a}, which the stations are able to lock and track throughout \threeD space.
      Due to the lightweight and miniature dimensions of the particular reflector, the stations were capable to provide only the \mbox{\threeD} position of the reflector relative to a coordinate system of the stations at frequency of \SI{5}{\Hz}.
      The reference attitude was determined offline by ICP algorithm with parameters set to maximize accuracy.
      During short occlusions between a station and the target, a predicted trajectory of the target is followed in order to focus back once the occlusions disappear.
      Hence, the data further used as a ground truth reference contain short time period outages as the stations initialized re-locking procedure.

\begin{figure}
      \captionsetup{width=0.95\columnwidth, captionskip=0pt}
      \centering
      \vspace*{-0.9em}
      \hspace*{-0.25em}
      \subfloat[\textit{Trial 1}: verification containing takeoff phase of the flight, where global estimation convergence and scan matching pipeline initialization is visible approx. at time \SI{8}{\second} and altitude of \SI{2.3}{\metre}
      ]{
        \includegraphics[width=0.99\columnwidth]{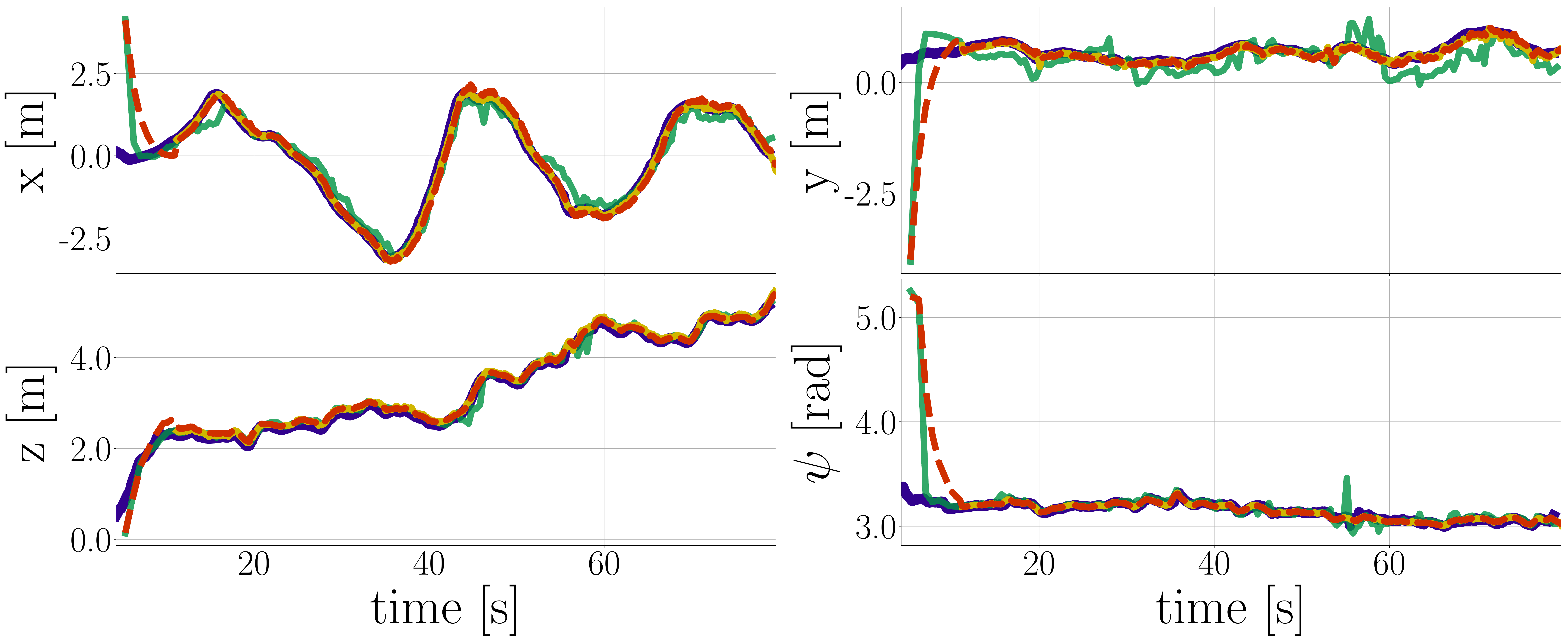}
        \label{fig:trial_2}
      }
      \vspace*{-0.6em}
      \subfloat[\textit{Trial 2}: verification containing ground truth reference interruptions around \SIlist{15;30;35}{\second} due to visual occlusions between the multi-stations and the tracked target
      ]{
        \includegraphics[width=0.99\columnwidth]{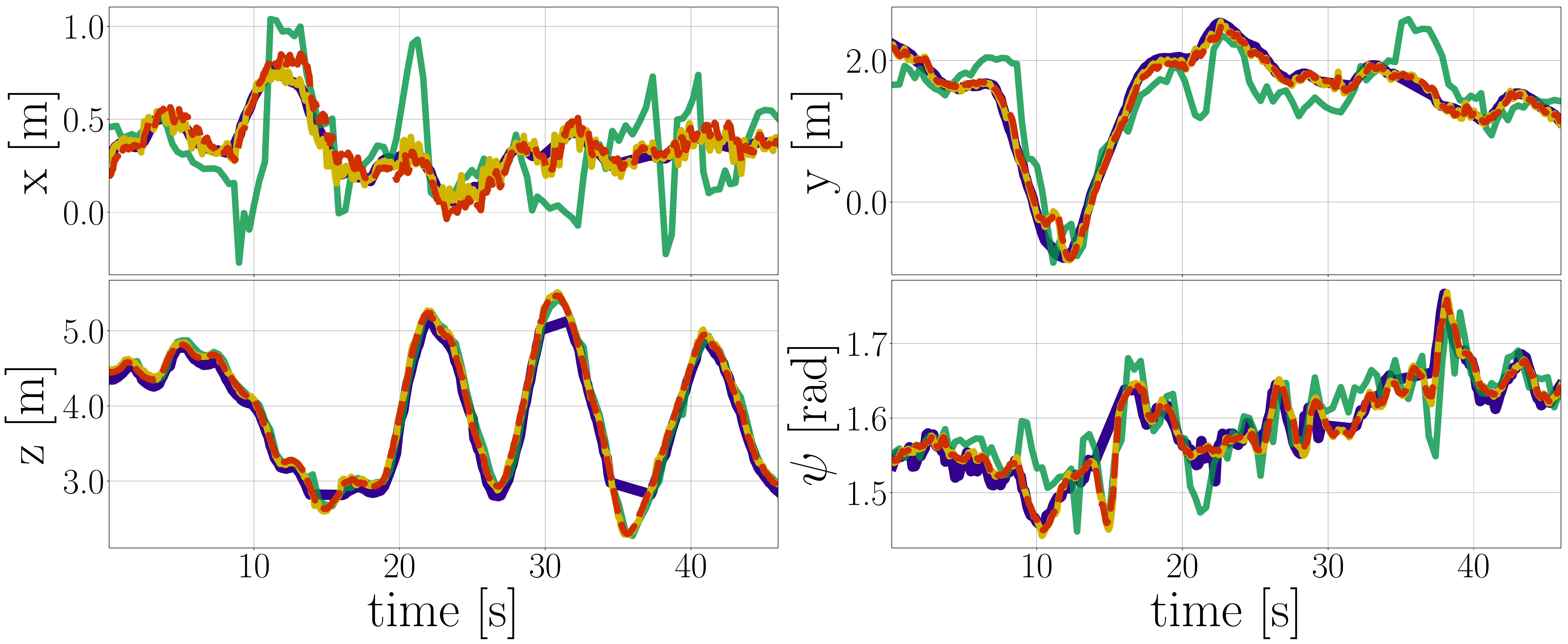}
        \label{fig:trial_1}
      }
      \vspace*{-0.2em}
      \caption{
        State variables $x, y, z, \psi$ for {\color[rgb]{0.22, 0.2, 0.502} ground truth} (\SI{5}{\Hz}), {\color[rgb]{0, .522, .243} global} (\SI{3}{\Hz}) and {\color[rgb]{.737,.667,0} local} (\SI{15}{\Hz}) localization, and {\color[rgb]{0.737,0.165,0} fused} state estimation (\SI{100}{\Hz}) during real deployment in church of St. Mary Magdalene in Chlum\'{i}n
      }
      \label{fig:quantitative_analysis}
      \vspace*{-1.8em}
    \end{figure}

    From multiple experimental flights tracked by an outer reference system, three particular trajectories are presented for which the quantitative results are given in \autoref{tab:experiments_accuracies}.
    Besides the table, outputs of the distinctive state estimation processes are outlined in \autoref{fig:quantitative_analysis} for the first two trials.
    The analysis of the localization system in real-world conditions exhibits estimation accuracy with translational RMSE less than \SI{0.23}{\metre} during each experiment.
    The experiments also demonstrate minimal time delay, smoothness, and robustness of the state estimation.
    These attributes are important for reliable deployment as their absence could lead to destabilization of the MAV control.
    The proposed localization system proves to be a reliable and robust source with sufficient precision of the position estimate.

    \sisetup{
      round-mode = places,
      round-precision = 3,
      table-parse-only,
      detect-weight=true,
      detect-inline-weight=text
    }
    \begin{table}[htb]
      \centering
      \begin{tabular}{lrrr}
        \toprule
        \multicolumn{1}{l}{Trajectory} & Trial 1 & Trial 2 & Trial 3\\\midrule
        length [\si{\metre}]                         & \num{24.0548} & \num{45.812} & \num{21.163} \\
        avg linear velocity [\si{\metre\per\second}] & \num{0.36058}  & \num{0.485} & \num{0.5051} \\
        max linear velocity [\si{\metre\per\second}] & \num{1.586}  & \num{2.294} & \num{1.734} \\


        \midrule
        RMSE translation [\si{\metre}] & \num{0.17919} & \num{0.1402} & \num{0.23026}\\
        RMSE absolute orientation [\si{\degree}]     & \num{2.381} & \num{2.4601} & \num{2.7468}\\
        max translation error [\si{\metre}] & \num{0.38537} & \num{0.5219} & \num{0.5944}\\
        max absolute orientation error [\si{\degree}]     & \num{6.8066} & \num{6.92827} & \num{11.3023}\\
        \bottomrule
      \end{tabular}
      \caption{Quantitative results of the \mbox{\threeD} position and absolute yaw orientation $\psi$ accuracy based on real data taken during real deployment in church of St. Mary Magdalene in Chlum\'{i}n
      }
      \label{tab:experiments_accuracies}
    \end{table}

    \hspace*{-1.2em}
    \begin{minipage}{.45\columnwidth}
      \centering
      \includegraphics[width=1\linewidth]{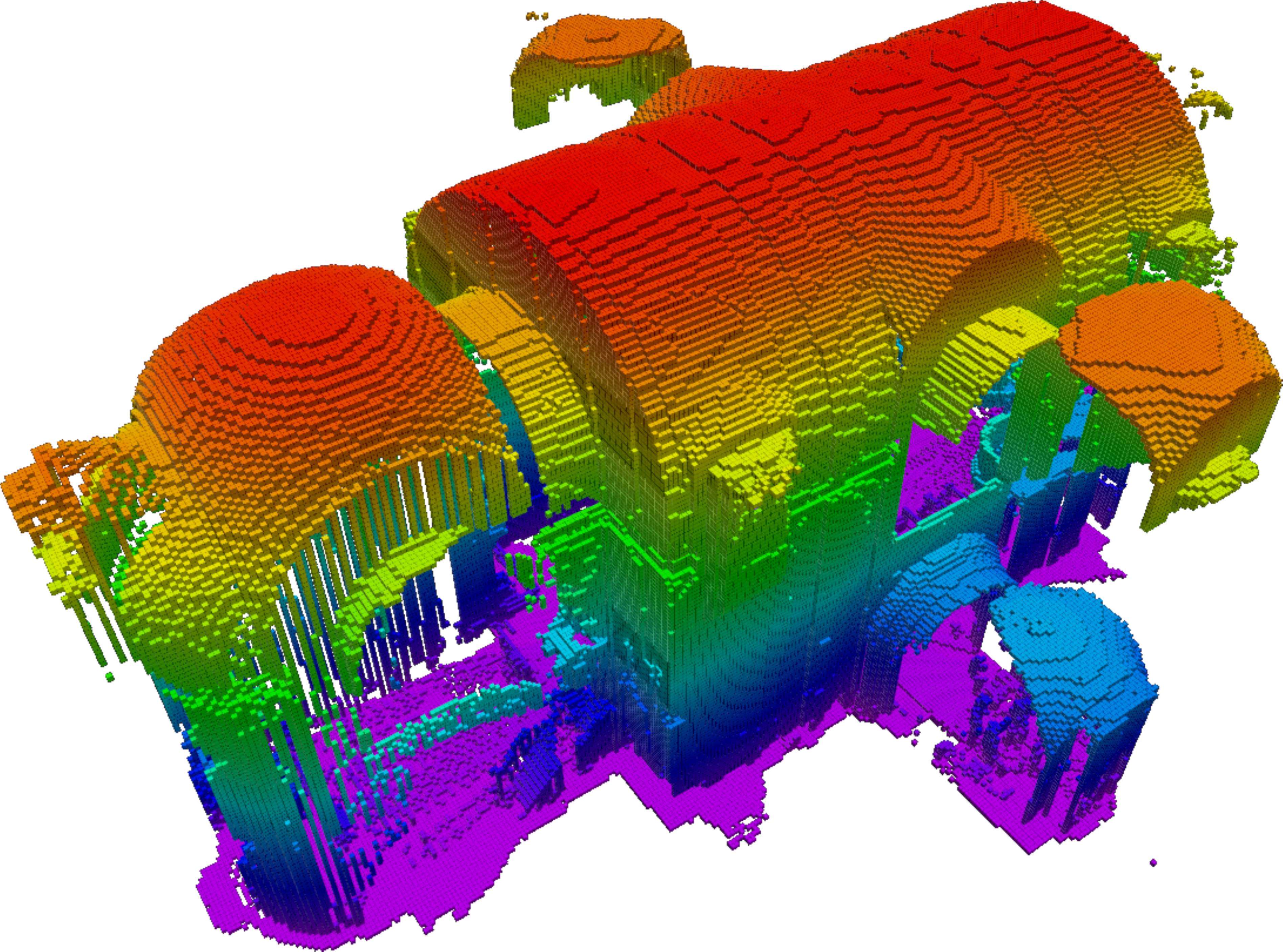}
      \captionsetup{width=0.99\linewidth}
      \captionof{figure}{Single-scan octree map of church of St. Mary Magdalene in Chlum\'{i}n}
      \label{fig:chlumin_octree}
    \end{minipage}%
    \hspace*{0.6em}
    \begin{minipage}{.48\columnwidth}
      \centering
      \includegraphics[width=1\linewidth]{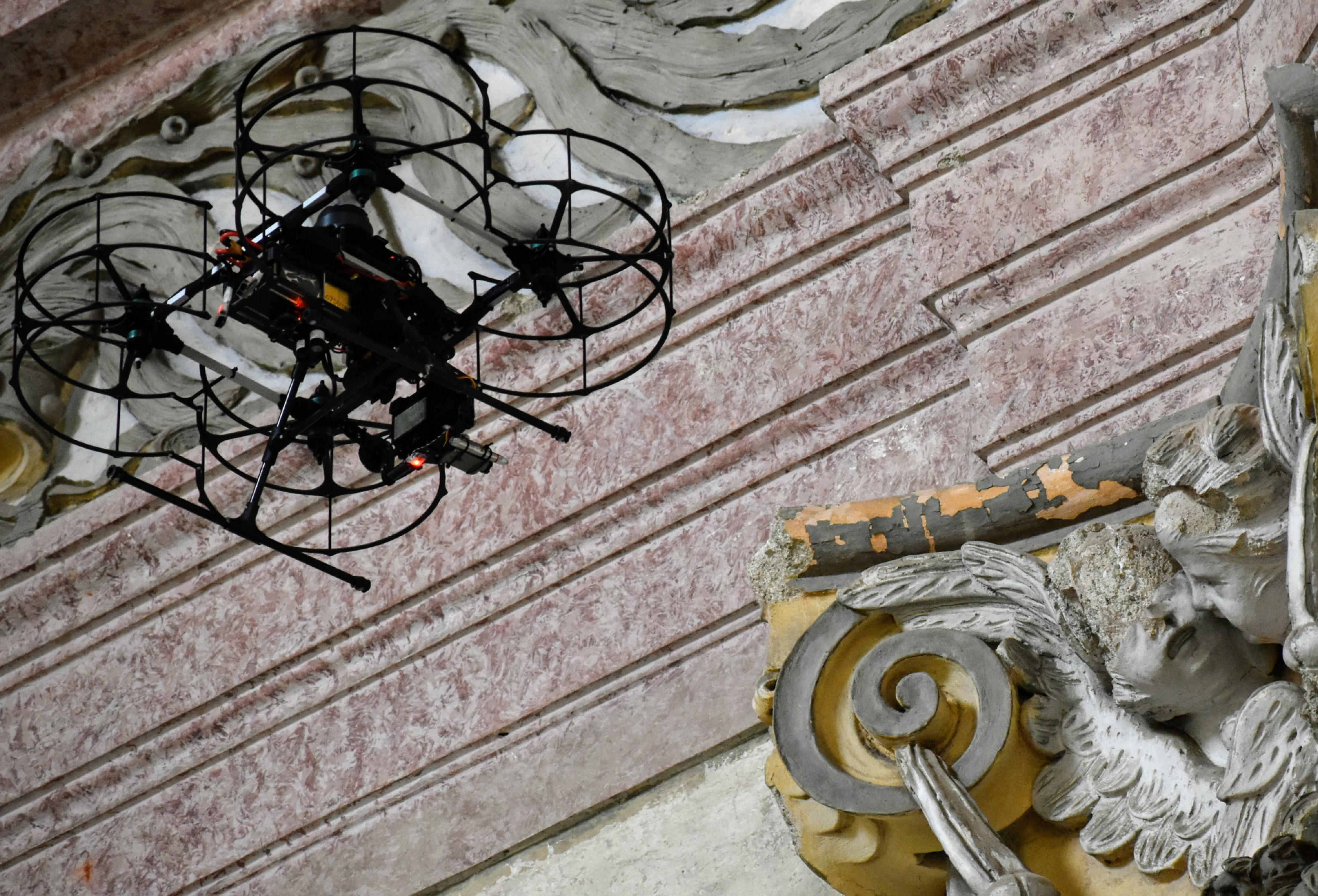}
      \captionsetup{width=0.99\linewidth}
      \captionof{figure}{Documentation in church of \mbox{St. Anne} and \mbox{St. James} in Star\'{a} Voda}
      \label{fig:stara_voda}
    \end{minipage}


    \vspace*{-0.4em}
    \section{Conclusion}
    \label{sec:conclusion}
    \vspace*{-0.05em}
    This letter presents the first comprehensive study on the use of autonomous MAV systems as an assistive technology for documentation of historical structures.
    The study shares the experience obtained during developing of the technology in close cooperation with team of restorers and conservationists, and discusses challenges of a robotic deployment.
    The proposed approach is validated and tuned on a set of identified tasks through extensive experimental flights aimed at collecting of exploitable data from the end-users.

    To provide state estimate in \textit{GNSS-denied} environments, an application-tailored localization system is presented, which was identified as the most important and challenging task in this application.
    This system provides local \threeD position and attitude without access to GNSS, and with the use of laser-inertial sensory setup copes with bad lighting conditions.
    That makes it feasible for deployment in indoor areas high above the ground, which are characteristic for historical monuments.
    The presented analysis of the localization system proves it to be a reliable and robust source of information with sufficient precision, which \mbox{enabled its deployment} into the feedback loop of the position control system.


    \bibliographystyle{IEEEtran}
    \bibliography{paper}

    \end{document}